\pdfoutput=1

\documentclass[11pt]{article}

\usepackage{ACL2023}
\usepackage{times}
\usepackage{latexsym}
\usepackage{url}            
\usepackage{booktabs}       
\usepackage{amsfonts}       
\usepackage{nicefrac}       
\usepackage{microtype}      
\usepackage{latexsym}
\usepackage{amssymb}
\usepackage{amsmath}
\usepackage{booktabs}
\usepackage{enumerate}
\usepackage{graphicx}
\usepackage{subfigure}
\usepackage{xspace}
\usepackage{float}
\usepackage{bbm}
\usepackage{bm}
\usepackage{multirow}
\usepackage{booktabs}
\usepackage{color}
\usepackage{framed}
\usepackage{stfloats}
\usepackage{iitem}
\usepackage{amssymb}
\usepackage{pifont}
\usepackage{arydshln}
\usepackage{enumitem}
\usepackage{wrapfig}
\usepackage{algorithm}
\usepackage{algpseudocode}

\usepackage[T1]{fontenc}
\usepackage{colortbl}
\definecolor{maroon}{cmyk}{0,0.87,0.68,0.32}

\newcommand{\gray}{\rowcolor[gray]{.90}}
\newcommand{\cellgray}{\cellcolor{gray!25}}

\usepackage[utf8]{inputenc}

\usepackage{microtype}

\usepackage{inconsolata}
\usepackage{multirow}

\title{CommonIT: Commonality-Aware Instruction Tuning for \\ Large Language Models via Data Partitions}

\author{
Jun Rao$^{1}$~~
 Xuebo Liu$^{1}$\thanks{~~Corresponding Author}~~
 Lian Lian$^{2}$~~
 Shengjun Cheng$^{2}$~~
    \bf{Yunjie Liao}$^{1}$~~
    \bf{Min Zhang}$^{1}$\\
    \textsuperscript{\rm1}Institute of Computing and Intelligence, Harbin Institute of Technology, Shenzhen, China \\
    \textsuperscript{\rm2}Huawei Cloud Computing Technologies Co., Ltd.~~~
    \\
    \texttt{\{rao7jun,yunjie445\}@gmail.com, \{liuxuebo,zhangmin2021\}@hit.edu.cn}\\
    \texttt{\{lianlian3,chengshengjun\}@huawei.com}
    }

\begin{document}
\maketitle
\begin{abstract}
With instruction tuning, Large Language Models (LLMs) can enhance their ability to adhere to commands. 
Diverging from most works focusing on data mixing, our study concentrates on enhancing the model's capabilities from the perspective of data sampling during training. 
Drawing inspiration from the human learning process, where it is generally easier to master solutions to similar topics through focused practice on a single type of topic, we introduce a novel instruction tuning strategy termed CommonIT: {C}ommonality-aware {I}nstruction  {T}uning.
Specifically, we cluster instruction datasets into distinct groups with three proposed metrics (\textsc{Task}, \textsc{Embedding} and \textsc{Length}). 
We ensure each training mini-batch, or ``partition'', consists solely of data from a single group, which brings about both data randomness across mini-batches and intra-batch data similarity. 
Rigorous testing on LLaMa models demonstrates CommonIT's effectiveness in enhancing the instruction-following capabilities of LLMs through IT datasets (FLAN, CoT, and Alpaca) and models (LLaMa2-7B, Qwen2-7B, LLaMa 13B, and BLOOM 7B).
CommonIT consistently boosts an average improvement of 2.1\% on the general domain (i.e., the average score of Knowledge, Reasoning, Multilinguality and Coding) with the \textsc{Length} metric, and 5.2\% on the special domain (i.e., GSM, Openfunctions and Code) with the \textsc{Task} metric, and 3.8\% on the specific tasks (i.e., MMLU) with the \textsc{Embedding} metric. Code is available at \url{https://github.com/raojay7/CommonIT}.
\end{abstract}

\section{Introduction} 
The emergence of ChatGPT~\cite{gpt4} and a range of large language models (LLMs) ~\cite{llama,bloom} brings light to artificial general intelligence (AGI).
As the size of the model increases, instruction fine-tuning becomes necessary to align human intentions with machine understanding of language. 
Compared to traditional fine-tuning, instruction tuning (IT) represents different tasks through instructions followed by some task-specific inputs~\cite{iyer2022opt}. 
Current researchers have obtained good results by passing only the IT stage~\cite{zhang-etal-2024-two} on datasets generated by multiple construction methods ~\cite{alpaca,vicuna,xu2023baize,fan2024reformatted}.
Typically after this stage of training, LLMs show a strong multi-task capability and the models can perform multifaceted tasks such as summarization, conversation, writing and other basic tasks.
\begin{figure}[t!]
    \centering
\includegraphics[width=1.0\linewidth]{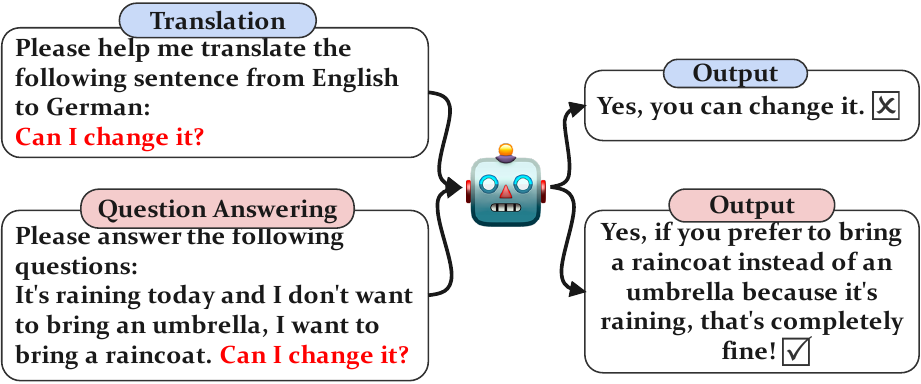}
    \caption{Because of the mix of instructions, the LLM cannot understand the specific task of the different instructions well after IT~\cite{shi2023specialist}. It fails to recognize the instructions of the translation task in this case, ``Please translate the following sentence ...'', and simply replies with the final phrase ``Can I change it?''.}
    \label{example}
\end{figure}

Existing research has already underscored the significance of data mixing in fine-tuning large models for achieving a generalized model. 
For example, \citet{flan_v2} pointed out that the key for LLMs in the IT phase is to enhance the understanding of instructions and they increase the diversity of instructions, e.g., by using more instructions to describe the same task~\cite{longpre2023flan}.
Despite this, these LLMs are susceptible to bias in the model's understanding of a particular task~\cite{kim2023taskweb} due to the mixing of multiple tasks or diverse instructions, which can lead to a decline in the average ability across multiple tasks~\cite{scialom-etal-2022-fine,iyer2022opt,wang2023far}. 
As illustrated in Figure \ref{example}, this situation results in a diminished ability of the model to follow instructions accurately. Specifically, it struggles to comprehend the requirements laid out in the instructions, leading to incorrect responses.

To enhance the model's ability to follow instructions, we introduce CommonIT, a methodology that enhances model comprehension of data features by engaging distinct data classes in individual gradient updates and interchanging data classes across batches. 
Central to our approach is the principle of data commonality~\cite{cui2022zerovl}, inspired by humans' learning process when preparing for exams. This means focusing preparation on one exam subject at a time rather than attempting to study for multiple subjects simultaneously.
Specifically, we advocate a two-phase fine-tuning process for LLMs. Initially, raw data are segmented into several groups and partitioned into designated sizes for training. 
This strategy enhances the model's ability to follow instructions, as evidenced by comprehensive evaluations across four dimensions of capacity testing.

Our experiments demonstrate its applicability across diverse IT datasets and potential for future applications in different models and domain-specific tasks.
Our findings indicate that length serves as the most effective criterion for grouping within the generic domain. 
Furthermore, embedding divisions should be tailored specifically for each task, and task-related information is crucial for optimizing fine-tuning quality in specific domains.
Our analysis of commonalities, from the perspectives of data quality, input question sentence representation and multi-task generalizability, delves into the reasons for improvements, demonstrating that the common learning strategy indeed enhances the effectiveness of representations and the capabilities across various sub-tasks.

Our contributions are as follows:
\begin{itemize}
\item  We propose the CommonIT framework, which leverages commonalities to enhance models' capabilities in following instructions. This framework includes three strategies for categorizing groups within IT datasets and incorporates a batch-based constraint policy for optimization. (\S \ref{sec:method}).
\item CommonIT demonstrates broad applicability across multiple dimensions, including various datasets, general and specialized domains, and diverse models. Additionally, we have explored scenarios to determine the most suitable group strategy for each context
(\S \ref{main result}).

\item Our exploration of commonalities provides a possible explanation for the sources of improvements 
(\S \ref{ablation} and \S \ref{sec:analysis}).

\end{itemize}

\section{Related Work}




\subsection{Data-centric AI}
There are current discussions on the curriculum~\cite{bengio2009curriculum, platanios2019competence,feng2023citing,lee2023instruction}, data mixing~\cite{wang2023far,xu2023baize}, and data filtering~\cite{zhou2023lima,anonymous2024alpagasus,xie2023data} during the training of large language models. 
LIMA~\cite{zhou2023lima} finds better performance could be achieved with just a few percent of samples from these datasets if selected properly with the very large models (65B). 
When smaller scale models are used (7B), the amount of data in SFT is usually related to the model size. Alpaca~\cite{alpaca} demonstrates that models at the 7B level, fine-tuned with a small amount of data (less than 100K), can exhibit strong alignment capabilities.
Further, AlpaGasus~\cite{anonymous2024alpagasus} found that models at the 7B level can also achieve strong alignment capabilities with fewer data. 
We have initiated early explorations in the data sampling strategy and find that the training scheme possibly has a major impact on the model's final performance under the small training epochs.

\subsection{Instruction Tuning}
%


Many works~\cite{alpaca,alpaca-gpt4,xu2023baize,phoenix-2023,wang2023far} employ distilled datasets and curate various data for fine-tuning language models, achieving enhanced performance.
Early works like FLAN 2021~\cite{flan2021} and Super-Natural Instructions~\cite{supernli} are to convert traditional NLP tasks into instruction format through manually defined instruction templates. 
FLAN-CoT~\cite{cot} and FLAN 2022 \cite{flan_v2} employ Chain-of-Thought training prompts to strengthen the reasoning of the model.
\citet{flan-moe} state that instruction finetuning can be considered a ``continual finetuning stage''.
\citet{DBLP:conf/acl/SuSKWHOYSZ023} demonstrate how various text encodings represent different tasks, enabling a single model to accomplish multiple downstream tasks and thereby achieving a more generalized model. 
These works have pointed out that the variety and scale of instruction data design tasks can significantly affect the ability of a model to generalize. 
However, existing efforts have primarily concentrated on exploring the data mixing~\cite{longpre2023flan,iyer2022opt}. 
In contrast, our work focuses on the training methodology using the same data, where we conduct a detailed analysis. 
By examining the various strategies and techniques in the training process, we aim to uncover insights that may transcend mere data considerations, potentially leading to more effective models that are attuned to the intricacies of instruction tuning.

\begin{figure*}[t]
    \centering
    \includegraphics[width=\linewidth]{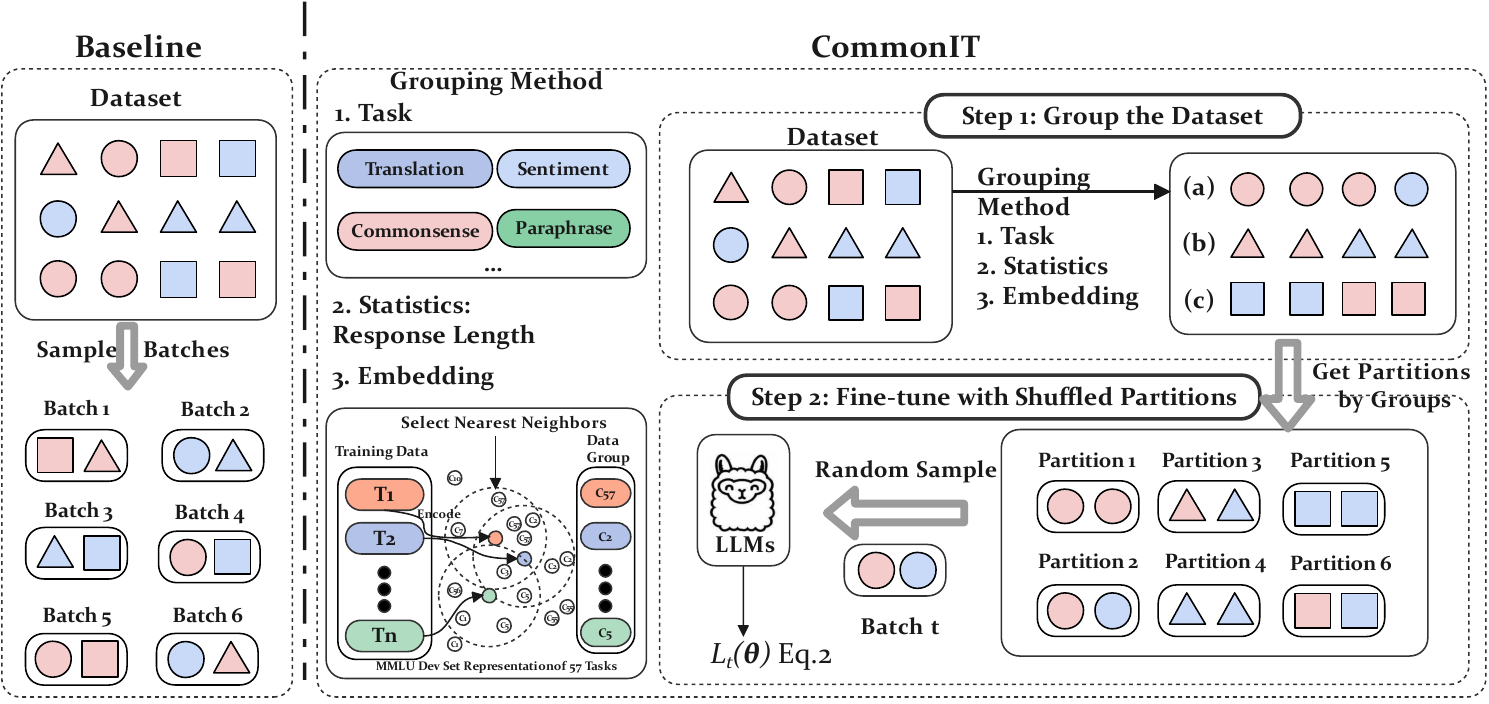}
    \caption{An overview of the baseline (IT) and our CommonIT. The different \textbf{shapes} and \textbf{colors} in the figure indicate a property of the data that can be used for grouping (task, statistics and embedding), and we use shapes as an example here. The CommonIT strategy inputs training data to the model during training from a group, e.g., batch $t$ from class (a). The model calculates the loss ${L}_t(\boldsymbol{\theta})$ of this partitioned data to update the parameters of the model.}
    \label{overview}
\end{figure*}

\section{Our Proposed CommonIT Method}\label{sec:method}
\textbf{Motivation} 
Data imbalance can lead to inconsistent performance across tasks, with some exhibiting exceptional results and others underperforming significantly~\cite{shi2023specialist}.
This phenomenon is typically evidenced by instruction misinterpretation, as illustrated in Figure \ref{example}, where the model fails to comprehend the user's intention accurately.
Our approach, CommonIT, aims to balance task-specific sampling for gradient updates with diverse data grouping.
This strategy draws upon the human learning process of understanding through analogy, leveraging similarities across different contexts to facilitate comprehension.
Through this approach, the model gains an enhanced ability to differentiate among the instructions of various tasks. Consequently, it can more accurately respond to distinct task directives, thereby minimizing the misinterpretation of instructions and reducing the generation of irrelevant content (\S \ref{exp}).

\paragraph{Overview}

As shown in Figure \ref{overview}, CommonIT is divided into two steps. 
First, we need to perform a clustering operation to categorize the dataset into multiple categories. 
It is worth noting that the clustering here does not require precise categorization but only focuses on a certain aspect of the dataset division. 
In the figure, the square, circle, and triangle indicate three kinds of data with different attributes. 
At the same time, the different colors indicate that the data can also be divided by the attribute of color, which is not divided here just due to space constraints. 
The second step is to construct partitions of the divided data by batch size and randomly take the divided partitions as one batch when the data are fed into the model.

\subsection{Background}\label{sft}
After large-scale pre-training, instruction tuning is the next phase of LLMs to enable the model to understand and interpret instructions for human language preferences~\cite{zhang2022opt,bloom,llama}. 

For a training data source $\mathcal{D}=\left\{\boldsymbol{x}^n, \boldsymbol{y}^n\right\}$, 
the standard instruction tuning of the language model is trained with maximum likelihood estimation: 
\begin{equation}\label{eq:mle}
\begin{aligned}
\hat{\boldsymbol{\theta}} & = \underset{\boldsymbol{\theta}}{\arg \max } \sum_{n=1}^N \sum_{m=1}^M \log P\left(\boldsymbol{y^n_m} \mid \boldsymbol{y^n_{<m}} ; \boldsymbol{x^n} ; \boldsymbol{s^n}; \boldsymbol{\theta}\right)
\end{aligned}
\end{equation}

where $\boldsymbol{s}, \boldsymbol{x}$ and $\boldsymbol{y}$ represent the instruction, the input, and the target, respectively. The $\boldsymbol{\theta}$ are the parameters to be optimized during the language model training and $M$ is the sequence length of the target.
Generally, the instructions are required and represent a specific task. Here is an example of an instruction: ``Please help me write a poem,'' requiring models to output content about poetry.
\subsection{Group the Dataset (GD)}\label{sec:gd}

The primary objective in organizing a dataset is to segregate distinct data categories, typically achieved by partitioning the data into tasks. 
Formally, we aim to decompose a dataset $\mathcal{D}$ into distinct sub-datasets $\mathcal{D}_0,\mathcal{D}_1,..., \mathcal{D}_n$. If task-based partitioning proves infeasible, data clustering methods can be employed for learning. 
Drawing an analogy with human learning, individuals often find it easier to grasp concepts when exposed to related topics concurrently, leveraging inherent similarities. 
Conversely, disparate topics can pose learning challenges.
We subsequently detail three potential group strategies:

\begin{itemize}
    \item \textbf{Group by Task} 
    The task-specific information facilitates the model's ability to differentiate between various task instructions, enhancing its generalization capacity across different instructional contexts.
    For the dataset that transforms traditional NLP tasks by designing instructions, we can obtain the task type of the original data. With this type, we can divide the dataset to get different categories. 
    
    \item \textbf{Group by Embedding}
    This approach is an alternative to embedding clustering, typically used when specific task categories cannot be readily identified. Automatic division can significantly enhance the effectiveness within specific domains for particular tasks.
    We use the category with the highest number of corresponding categories among the $k$ retrieved pieces of data as the category of the training set data. 
    For simplicity, we use the category of MMLU (development set corresponds to a category of 57 exams) to categorize the training data.
    We first use a certain sentence encoder to convert sources in both the training set and development set to vector representations.\footnote{https://huggingface.co/sentence-transformers/all-mpnet-base-v2} 
    Then, for each training source $\boldsymbol{s}$, we retrieve its nearest $k$ neighbors $\boldsymbol{s}_1, \boldsymbol{s}_2, ..., \boldsymbol{s}_k$ from the MMLU development set (according to the distances in the sentence encoder’s embedding space).~\footnote{https://github.com/Shark-NLP/self-adaptive-ICL}
   Given some predefined similarity measure $d$, such as the cosine similarity, the neighbors are ordered in such a way that $d(\boldsymbol{s}_i, \boldsymbol{s}) \leq d(\boldsymbol{s}_j, \boldsymbol{s})$ when $i < j$.

    \item \textbf{Group by Statistics}
    Due to the nature of IT data, responses of the same length typically belong to the same data category, such as multiple-choice, question-answering, or translation tasks. This inherently includes task similarity within the data. 
    The length can be the simplest metric for classifying clusters when the task information is missing.
    We count the length distribution of each IT dataset such that the number of samples in the sub-datasets after each IT dataset is divided remains the same for each length interval.
\end{itemize}

\subsection{Fine-tune with Shuffled Partitions (FP)}

In conventional model training, the batch data is usually sampled randomly from the entire data source $\mathcal{D}$. 
The training of language model employs \textit{mini-batch gradient descent} rather than \textit{batch gradient descent} or \textit{stochastic gradient descent}.

CommonIT supposes that the mini-batches are bucketed
in a particular way and upgrade these samples.
In our proposed CommonIT, we can perform the batch construction as follows since we obtain the ``class division'' of the data.
We ensure instances within each batch come from the same group while the order of sampled groups is random. 
Take Figure \ref{overview} for example, the first batch $\mathcal{B}^*_1$ comes from a group (a), and the $t$th batch $\mathcal{B}^*_t$ may be composed of samples of the only group (a) or (b), or (c).
So the batch loss can be calculated as:
\begin{equation}\label{eq:loss}
L_t(\boldsymbol{\theta}) = -\frac{1}{N} \sum_{i=1}^{N} \log P(\boldsymbol{y}^{(i)}_t|\boldsymbol{x}^{(i)}_t;\boldsymbol{s}^{(i)}_t;\boldsymbol{\theta})
\end{equation}
where $\boldsymbol{s}, \boldsymbol{x}$ and $\boldsymbol{y}$ represent the instruction, the input, and the target, respectively. The $L_t(\boldsymbol{\theta})$ is the training loss of a mini-batch that has $N$ examples at the $t$-th training step.

\section{Evaluation Setup}
\subsection{Evaluations}
Factual knowledge, reasoning, multilinguality, and coding are foundational to LLMs, as they encapsulate the core capabilities required to comprehend, analyze, and generate human-like text. 
In the evaluations, we examine the model's performance in these areas, employing four out-domain benchmarks to assess its ability to assimilate knowledge (MMLU~\cite{MMLU}) and execute complex reasoning (BBH~\cite{bbh}), multilinguality (TydiQA~\cite{tydiqa}), and coding (Codex-Eval~\cite{humaneval}) tasks. 
\begin{itemize}
    \item \textbf{Factual Knowledge} represents a critical dimension of capability in LLMs, essentially reflecting their memory capacity. 
    We employ the Massive Multitask Language Understanding dataset (MMLU~\cite{MMLU} ) as a benchmark to measure the model's factual knowledge. 
    \item \textbf{Reasoning} is another crucial capability for large models, particularly in solving complex problems. 
    We utilize the Big-Bench-Hard dataset (BBH~\cite{bbh}), comprising 23 intricate tasks, to assess the model's general reasoning capabilities.
    \item \textbf{Multilinguality} is essential for enabling large models to serve speakers of various languages. 
    We employ the TyDiQA~\cite{tydiqa} dataset, a multilingual testing dataset encompassing 11 different language types.
    \item \textbf{Coding} is another important ability that people need large language models. 
    We use the HumanEval dataset~\cite{humaneval} (we refer to it as Codex-Eval) to evaluate the models’ capability to generate functionally correct programs from docstrings.
\end{itemize}

\subsection{Training Datasets}
We have selected several representative IT datasets from distinct sources, encompassing task-constructed data and model-generated high-quality data. 
Our diverse selection facilitates a thorough evaluation, enabling us to gauge performance across varied data types and characteristics. The selections include:
\textbf{FLAN V2}~\cite{flan_v2} A collection of NLP tasks that combines several existing datasets with various data augmentations.
\textbf{FLAN CoT}~\cite{cot} A collection of datasets annotated with chain-of-thoughts. We use the CoT mixture from the FLAN v2 collection~\cite{flan_v2}, splitting it out as a separate dataset.
\textbf{Alpaca GPT4}~\cite{alpaca-gpt4} A dataset was created using the Alpaca dataset as inputs, replacing the example generations with generations from GPT-4.
Detailed information regarding the datasets and training settings can be found in Appendix~\ref{sec:appendix}.

\begin{table*}[t!]
\centering
\scalebox{0.8}
{
\begin{tabular}{{clccccccl}}
\toprule
\multirow{2}{*}{\textbf{Dataset}}&\multirow{2}{*}{\textbf{Model}}& \multicolumn{2}{c}{\textbf{MMLU}}  &  
\multicolumn{2}{c}{\textbf{BBH}}&\textbf{TydiQA}&\begin{tabular}[c]{@{}c@{}}\textbf{Codex-Eval}\end{tabular} &\multirow{2}{*}{\textbf{AVG.}}\\ \cmidrule(lr){3-4} \cmidrule(lr){5-6} 
\cmidrule(lr){7-7} \cmidrule(lr){8-8} 
                      & & \textbf{0-shot} &\textbf{5-shot}&\textbf{Direct}&\textbf{CoT} &\textbf{F1}&\textbf{P@10 }             \\ \midrule
-&LLaMa 7B
          &                   32.1&35.2  & 34.0&33.3&37.0&18.3&31.7 \\\hline
          \multirow{5}{*}{\textbf{FLAN CoT}}
          &IT* & 41.3&42.5 & 33.7&31.3 & 44.4&17.3&35.1       \\
          &IT& 37.1&38.3 & 32.9&34.1 & 47.5&19.3&34.9       
                   \\
          &CommonIT &&&&&&\\
          & \cellgray\quad By Embedding  &\cellgray    \textbf{40.2}&\cellgray41.4  &\cellgray\textbf{36.1}&\cellgray33.5 &\cellgray45.8&\cellgray17.4&\cellgray35.7 \\
          & \cellgray\quad By Length  &    \cellgray38.7&\cellgray\textbf{42.3}  &\cellgray33.4&\cellgray\textbf{35.2} &\cellgray\textbf{47.9}&\cellgray\textbf{20.2}&\cellgray\textbf{36.3} 
\\ \hline
\multirow{5}{*}{\textbf{Alpaca}}
&IT*          & 42.6&38.3 & 28.5&32.3&23.6&25.0 &31.7              \\
&IT           & 34.8&36.4 & 32.6&33.0&37.4&22.6 &32.8              \\
&CommonIT &&&&&&\\
&  \cellgray\quad By Embedding              &\cellgray  \textbf{41.1}&\cellgray40.1 &\cellgray \textbf{33.6}&\cellgray33.8\cellgray&\cellgray38.7&\cellgray23.0& \cellgray35.1 \\
&  \cellgray\quad By Length              &\cellgray  40.4&\cellgray\textbf{40.1} &\cellgray 33.5&\cellgray\textbf{34.6}&\cellgray\textbf{38.9}&\cellgray\textbf{24.7}&\cellgray\textbf{35.4}  \\\hline
\multirow{6}{*}{\textbf{FLAN}}
&IT*& 45.4&47.1 & 38.6&36.1&45.0&12.9& 37.5  \\
&IT& 44.2&45.2 & 38.3&37.2&45.1&16.8& 37.8   \\ 
&CommonIT &&&&&&\\
&  \cellgray\quad By Task & \cellgray{46.6}&\cellgray{47.4} &           \cellgray{38.9}&\cellgray{37.2}&\cellgray{45.7}&\cellgray{19.6}&\cellgray{39.2} \\
&  \cellgray\quad By Embedding & \cellgray\textbf{47.2}&\cellgray\textbf{48.5} & \cellgray{38.9}&\cellgray{37.9}&\cellgray{44.5}&\cellgray\textbf{21.8}&\cellgray{39.8} \\
& \cellgray\quad By Length &\cellgray {46.7}&\cellgray{47.9} &\cellgray           \textbf{39.7}&\cellgray\textbf{39.9}&\cellgray\textbf{47.2}&\cellgray{19.3}&\cellgray\textbf{40.1} 
\\ \bottomrule
\end{tabular}
}
\caption{Main results for CommonIT at LLaMa-7B. 
``*'' denotes the results from  ~\citet{wang2023far}.
\textbf{The large model of multidimensional measurements shows the best results using the Length metric.}
}\label{tab:main}
\end{table*}

\begin{table*}[t!]
\centering
\scalebox{0.8}
{
\begin{tabular}{clccccc}
\toprule
\multirow{2}{*}{\textbf{Dataset}}&
\multirow{2}{*}{\textbf{Model}} & \multicolumn{5}{c}{\begin{tabular}[c]{@{}c@{}}\textbf{MMLU} \\ \textbf{(0-shot/5-shot)}\end{tabular}} \\ \cline{3-7} 
                      & & \textbf{Humanities} & \textbf{Social.} & \textbf{STEM}  & \textbf{Other} & \textbf{AVG.}           \\ \midrule
                      - & LLaMa 7B
          & 31.5/31.5      & 31.2/37.3   & 29.7/32.3 & 36.1/41.3                  & 32.1/35.2  \\\hline
                   \multirow{4}{*}{\textbf{FLAN CoT}}&IT                 & 34.6/{36.7}     & 41.1/42.3   & 30.8/31.1 & 42.7/43.5 & 37.1/38.3               \\
                   &CommonIT&&& \\
&\cellgray\quad By Length                  & \cellgray{34.9}/\textbf{39.2}           & \cellgray\cellgray\cellgray\textbf{44.4/48.2}        &\cellgray\cellgray  \textbf{32.7/33.9}     & \cellgray  44.5/\textbf{49.1}    &\cellgray   38.7/\textbf{42.3} \\
&\cellgray\quad By Embedding                  &\cellgray \textbf{38.3}/38.3           &\cellgray {43.5/46.1}        &\cellgray  {32.1/33.7}     & \cellgray  \textbf{47.3}/48.6    & \cellgray  \textbf{40.2}/41.4 
\\ \hline
\multirow{4}{*}{\textbf{Alpaca}}&
IT           &34.4/35.4     & 35.3/35.4  & 28.2/30.5 & 41.0/41.3 & 34.8/36.4              \\
&CommonIT&&& \\
& \cellgray\quad By Length                  & \cellgray38.3/\textbf{38.5}     &\cellgray {43.4/42.8}   &\cellgray {32.2/32.2} &\cellgray 48.3/\textbf{47.1} & \cellgray{40.4/40.1} \\
& \cellgray\quad By Embedding                  &\cellgray \textbf{39.6}/37.8     &\cellgray \textbf{44.1/43.2}   & \cellgray\textbf{32.5/33.3} & \cellgray\textbf{48.5}/46.9 &\cellgray \textbf{41.1/40.1} 
\\ \hline
\multirow{5}{*}{\textbf{FLAN}}&IT                     & 42.7/42.1   & 49.9/50.7 & 34.1/37.0   & 50.5/52.1 & 44.2/45.2        \\
&CommonIT&&& \\
& \cellgray\quad By Task     &\cellgray {44.0/45.2}      & \cellgray{53.8/54.3}  & \cellgray{36.9/38.0} & \cellgray{52.5/52.6} & \cellgray{46.6/47.4} \\
& \cellgray\cellgray\quad By Length     &\cellgray {44.5/44.5}      & \cellgray{53.6/54.7}  &\cellgray {35.9/39.5} &\cellgray {53.3/54.2} &\cellgray {46.7/47.9} \\
& \cellgray\quad By Embedding     & \cellgray\textbf{44.8/45.3}      & \cellgray\cellgray\textbf{53.9/55.6}  & \cellgray\textbf{37.3/39.8} & \cellgray\textbf{53.7/54.7} & \cellgray\textbf{47.2/48.5} 
\\ \bottomrule
\end{tabular}
}
\caption{The results of the CommonIT method across multiple task categories in MMLU. Using MMLU's sub-disciplines as discrete embedding divisions has enhanced the performance within each sub-discipline of the MMLU task, thereby improving the overall results. \textbf{The embedding is more effective than other metrics in MMLU.}
}\label{tab:forgetting}

\end{table*}
\begin{table}[t!]
\centering
\scalebox{0.775}
{
\begin{tabular}{llll}
\toprule
\textbf{Model/Domain} & \textbf{GSM} & \textbf{OpenFunctions} & \textbf{Code} \\
\midrule
{IT} & 39.0 & 30.4 & 23.6 \\
{Common IT}&&&\\
\gray \quad By Length& 36.0 (-3.0) & 31.3 (+0.9) & 28.2 (+4.6) \\
\gray \quad By Embedding & 39.0 (+0.0) & 34.8 (+4.4) & 28.3 (+4.7)\\
\gray \quad By Task & \textbf{43.5 (+4.5)} & \textbf{35.7 (+5.3)} & \textbf{29.3 (+5.7)} \\
\bottomrule
\end{tabular}
}
\caption{Comparison of methods across domain-specific IT. \textbf{Task metric in CommonIT is the key to improving domain-specific instruction following abilities.
}}\label{tab:domain_it}
\end{table}
\section{Experiment}\label{exp}

\subsection{Main Results}\label{main result}
\paragraph{CommonIT Improves Instruction Following In Several Dimensions} Table \ref{tab:main} illustrates the comparative results of the IT on LLaMa 7B of general IT datasets.
For LLaMa, the conventional IT approach (IT) has enhanced the model's performance across three capability dimensions, with an average improvement of 10 points in MMLU, 4 points in BBH, and 5 points in TydiQA. 
Employing our training strategy (CommonIT), we improved further on an already strong and influential baseline~\cite{wang2023far}.
This indicates CommonIT's effectiveness with varied data distributions.

\paragraph{\textsc{Length} Metric Works Best in General Domain}
Table \ref{tab:main} also presents the results of different grouping methods by combining different grouping strategies with three datasets. 
Since only the FLAN dataset contains partial task category information, we conducted ablation experiments on this dataset using three grouping methods. 
Due to the lack of original task divisions, we could only perform embedding and length ablation experiments in these two datasets.
Various grouping strategies improve the baseline results across multiple capability dimensions. 
These strategies yield varying degrees of enhancement in different capability areas. 
Employing embedding similarity as a group method also resulted in performance gains across these datasets.
As a simple grouping criterion, length has already achieved notable improvements. 
This significant gain demonstrates the importance of leveraging specific training approaches tailored to the unique properties of the data, ultimately leading to enhanced model performance on the IT datasets. 

\paragraph{\textsc{Embedding} Metric Works Best Based on the Specific Tasks}
Given that our experiments on embedding similarity are based on the division of MMLU's 57 categories, we further present detailed results for each task category in MMLU, as illustrated in Table \ref{tab:forgetting}.
The validation of exam questions across diverse subjects mirrors how humans prepare for exams in various disciplines.
Integrating our CommonIT method shows the model substantially enhances various disciplinary tasks relative to the baseline approach (IT). 
This indicates that training on similar questions grouped within a single batch using embedding can enhance the effectiveness of individual sub-tasks, thereby augmenting the overall capabilities. This observation further suggests that strategically designing the division of embeddings for specific tasks can enhance performance.

\paragraph{\textsc{Task} Metric Works Best in Special Domain}
To further demonstrate CommonIT's improved instruction-following abilities, we conducted instruction fine-tuning across specific domains, namely GSM~\cite{gsm}, OpenFunctions~\cite{patil2023gorilla}, and Code~\cite{wei2023magicoder}. 
We mix the dataset with the FLAN and the domain-specific data to establish a strong baseline (IT) using LLaMa-2 7B as the base model. 
For example, we mix the GSM 7.5K and FLAN 100K for training and evaluate the model for the GSM testing set.
The comparison methods (Length, Embedding, and Task) mixed the previously described grouped FLAN dataset by different group methods.

Our findings in Table \ref{tab:domain_it} indicate that, without further delicate domain-specific grouping, using task grouping directly improves the performance, proving CommonIT is enhanced by the Model's ability to discern the informational distinctions between tasks. 
Improvements in task-specific performance in domains suggest that the length metric may also carry domain-specific information beneficial to the model without requiring further processing, particularly in OpenFunctions and Code.
In contrast, embedding with MMLU grouping diminished performance in specific domains of fine-tuning, such as OpenFunctions and Code. 
These observations underscore the importance of constructing clear distinctions between task instructions and suggest that re-embedding for specific tasks could further optimize performance. Subsequent results, assessed across the four dimensions of competence, utilized length as the foundational criterion for division.



\begin{table}[t]
\centering
\scalebox{0.7}
{
\begin{tabular}{lccccl}
\toprule
\textbf{Model}         & \begin{tabular}[c]{@{}c@{}}\textbf{MMLU} \\ \textbf{0-shot}\end{tabular} & \begin{tabular}[c]{@{}c@{}}\textbf{BBH} \\ \textbf{Direct}\end{tabular}&\textbf{TydiQA}&\begin{tabular}[c]{@{}c@{}}\textbf{Codex}\\\textbf{Eval} \end{tabular}&\begin{tabular}[c]{@{}c@{}}\textbf{AVG. }\end{tabular}\\ \midrule
BLOOMZ 7B        

 & 42.1 & 28.1 &64.8&15.0 & 37.5                  \\
\quad+IT      
            
& 39.6  & 30.1 &69.7&17.1 &39.1 (+1.6)       \\
\gray\quad+CommonIT                      & \textbf{42.7} &\textbf{30.1}  &\textbf{71.6}&\textbf{18.3}&\textbf{40.7 (+3.2)}  \\  
\hline
LLaMa2-7B         & 36.6 &34.1  &49.1&  25.8  & 36.4                                         \\
\quad+IT      & 50.0    &40.0  &52.9&  20.7 & 40.9 (+4.5)                      \\
\gray\quad+CommonIT  & \textbf{50.8}  &\textbf{40.6}&\textbf{55.9} &\textbf{22.3}&\textbf{42.4 (+6.0)}\\
\hline
Qwen2-7B         & 68.3 &\textbf{49.0}  &  \textbf{27.1}  & 67.3 & 52.9                                       \\
\quad+IT      & 68.9    &30.6  &19.0&  79.0 & 49.4  (-3.5)                      \\
\gray\quad+CommonIT  & \textbf{69.1}  &47.2&26.0&\textbf{76.6}&\textbf{54.7 (+1.8)}

\\ \hline
LLaMa 13B         & 42.4 &38.7 &47.4&26.6 &38.8                                                                          \\
\quad+IT      & 46.1    &39.3 &51.2&27.2 &41.0 (+2.2)                                                                       \\
\gray\quad+CommonIT  &\textbf{47.1 }  &\textbf{39.5}&\textbf{52.4}&\textbf{31.3} & \textbf{42.6 (+3.8)}

\\ \bottomrule
\end{tabular}
}
\caption{Tuning results on FLAN for the different models with our CommonIT (Length).}\label{tab:scale}
\end{table}
\paragraph{Applicability across Various Foundation Models}
To further demonstrate the applicability of our approach across different models (architecture, scale and sequence length), we conducted experiments on the BLOOM 7B and LLaMa 13B and newly version of LLaMa (LLaMa2-7B). 
We evaluated the results of four benchmarks, as shown in Table \ref{tab:scale}.
CommonIT outperforms IT substantially, showcasing an average increase of about 1.5 points. 
The results indicate that our approach outperforms the baseline IT strategy, achieving a noteworthy improvement in various models. 

\subsection{Ablation Study}\label{ablation}
The success of CommonIT hinges on the initial grouping in Stage One (GD) and the subsequent assurance in Stage Two (FP) that data within a single batch exclusively originates from a single group. 
We conducted ablation studies on these two components and have presented the corresponding results (w/o GD and w/o FP) in Table \ref{tab:ablation}. 
Previous main results have demonstrated that the length metric is most effective in general domain IT, and we have adopted this metric as the foundational basis for CommonIT.
It is observable that in the absence of GD (baseline IT), all evaluation results experience a significant decline, underscoring the critical importance of grouping for the training process. 
Furthermore, the omission of FP 
led to a significant drop in most results. 

\begin{table}[t]
\centering
\scalebox{0.7}
{
\begin{tabular}{lccccc}
\toprule
\textbf{Model}         & \begin{tabular}[c]{@{}c@{}}\textbf{MMLU} \end{tabular} & \begin{tabular}[c]{@{}c@{}}\textbf{BBH} \\ \end{tabular}&\textbf{TydiQA}&\begin{tabular}[c]{@{}c@{}}\textbf{Codex}\\\textbf{Eval}\end{tabular}&{\textbf{AVG.}}\\ \midrule
LLaMa 7B&32.1&34.0&37.0&18.3&30.4\\ \hline
FLAN&{46.7}                                                           & {39.7} &45.2& 16.8 &37.1                                       \\
\gray  \quad w/o GD & 44.2                                                            & 38.3 &45.1&12.9&35.1\\ 
\gray \quad w/o FP    &44.4&36.0&38.5&16.8&33.9                                                         \\
\hline
FLAN CoT&{38.7}                                                           & 33.4 &47.9&20.2&35.1\\
\gray \quad w/o GD&    37.1&32.9&44.5&19.3 &33.5       \\    
\gray \quad w/o FP      &37.5&32.6&47.2&19.4 &34.2                                        
\\ \hline
Alpaca&40.4                                                           & 33.5&38.9& 24.7 &34.4   \\
\gray \quad w/o GD    & 34.8                                                            & 32.6&37.4&22.5&30.0\\               
\gray \quad w/o FP      &39.2&15.6&30.5& 28.0 &28.3                   
                           \\ \bottomrule
\end{tabular}
}
\caption{Ablation study on CommonIT (Length). ``w/o GD'': the baseline results of the IT. ``w/o FP'': the model undergoes sequential training on predefined groups.
}\label{tab:ablation}
\end{table}

\subsection{Analysis}
\label{sec:analysis}
\paragraph{Learning from High\&Low-Quality Instructions} We compare one existing state-of-the-art method: Alpagasus: GPT 3.5 selected the highest scoring 9,000 samples. We show the average results of multitasking with a uniform setup in Table \ref{tab:sota}, showing that CommonIT gets the best score (37.1).
To further ensure a fair comparison, we supplemented our evaluation with GPT assessments and results from Alpaca-Eval~\cite{dubois2024alpacafarm}.
CommonIT achieves better results in dialogue scenarios gained close to a 10-point boost, indicating that our method obtains more reliable and higher-quality replies.
Alpagasus~\cite{anonymous2024alpagasus} belongs to data filtering approaches, whereas ours is a data learning approach. 
They focus on learning from a subset of high-quality data, while CommonIT enables the model to learn from good and bad data. The results show that CommonIT demonstrates a superior understanding and learning of High\&Low-quality instructions.

\begin{table*}[ht!]
\centering
\scalebox{0.8}
{
\begin{tabular}{{lccccccccl}}
\toprule
\multirow{2}{*}{\textbf{Model}} &\multicolumn{2}{c}{\textbf{MMLU}}  &  
\multicolumn{2}{c}{\textbf{BBH}}&\multicolumn{1}{c}{\textbf{TydiQA}}&\begin{tabular}[c]{@{}c@{}}\textbf{Codex-Eval}\end{tabular}&\multirow{2}{*}{\begin{tabular}[c]{@{}c@{}}\textbf{Win Rate}\\\textbf{(Standard Error)}\end{tabular}}&\multirow{2}{*}{\textbf{AVG.}}\\ \cmidrule(lr){2-3} \cmidrule(lr){4-5} 
\cmidrule(lr){6-6} \cmidrule(lr){7-7} 
                       & \textbf{0-shot} &\textbf{5-shot}&\textbf{Direct}&\textbf{CoT} &\textbf{F1}&\textbf{P@10}         \\ \midrule   
Alpaca IT         & 34.8 &36.4&32.6&33.0&37.4&22.6&36.5\small{(1.7)} &33.3  \\
\begin{tabular}[c]{@{}c@{}}
 Alpagasus 
\small{~\cite{anonymous2024alpagasus}}\end{tabular} 
& 38.1&37.4                                                            & 31.6 &33.4&44.3&22.8&37.7\small{(1.7)}&35.0 (+1.7) \\ 
                           

\gray Alpaca CommonIT      & \textbf{40.4}&\textbf{40.1}                    & \textbf{33.5}&\textbf{34.6}&\textbf{38.9}& \textbf{24.7} &\textbf{47.3\small{(1.7)}} &  \textbf{37.1 (+3.8)}                                                      \\
\bottomrule
\end{tabular}
}
\caption{Comparison of state-of-the-art method (Alpagasus) focusing on high-quality data and IT using high\&low quality data on LLaMa 7B. 
We further compare Win Rate on AlpacaEval as evaluated by ChatGPT for evaluating open-ended intractability. CommonIT (Length) significantly outperforms the comparison methods in Win Rate, indicating that it effectively learned the difference between instructions for high and low-quality data.
}\label{tab:sota}
\end{table*}
\begin{figure*}[t]
\makeatletter
\renewcommand{\@thesubfigure}{\hskip\subfiglabelskip}
\makeatother
\centering
\subfigure[]{
\begin{minipage}[b]{0.25\linewidth}
    \centering
\includegraphics[width=1\linewidth]{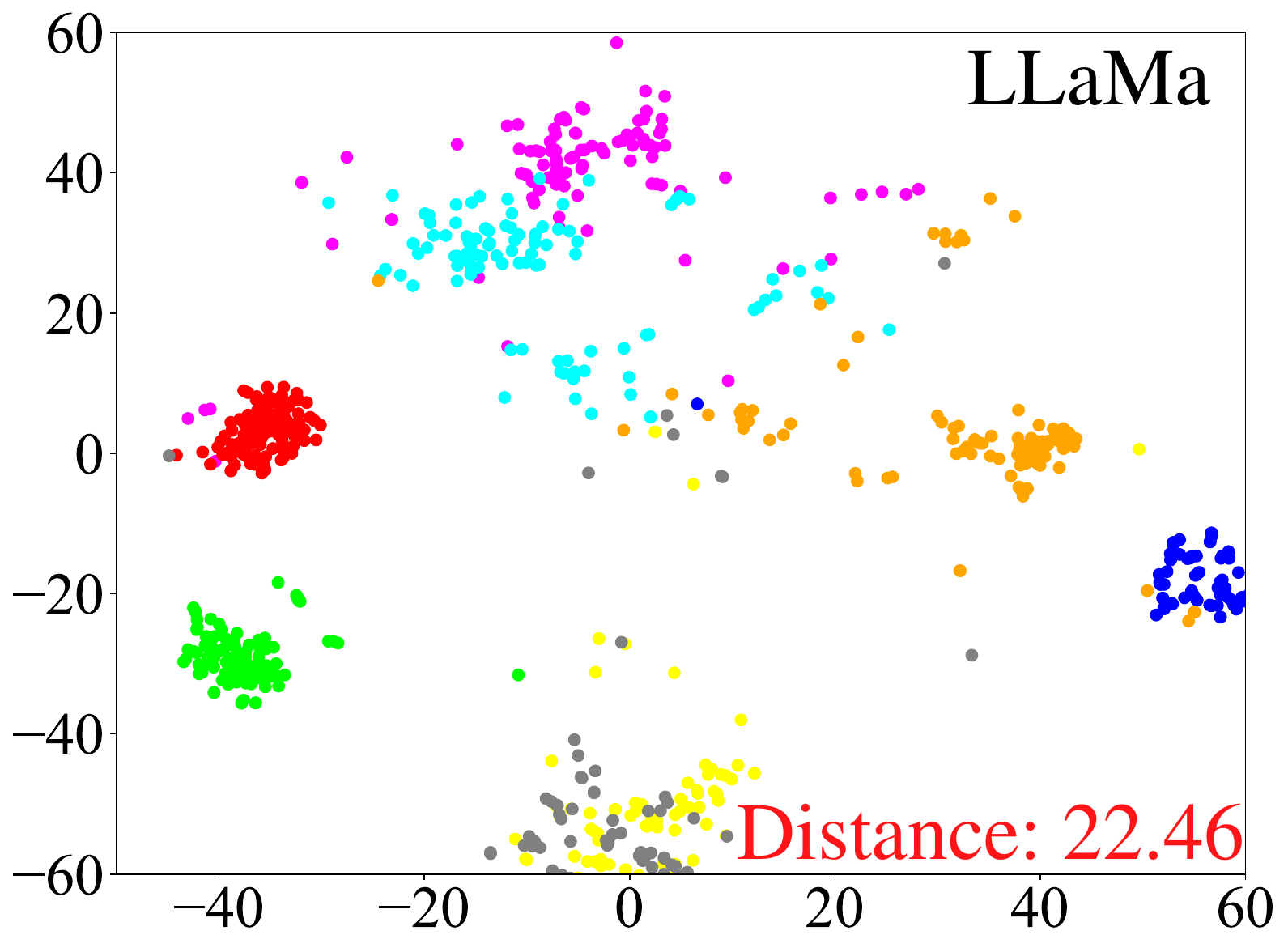}
    \label{vanilla}
\end{minipage}
}
\subfigure[]{
\begin{minipage}[b]{0.25\linewidth}
    \centering
\includegraphics[width=1\linewidth]{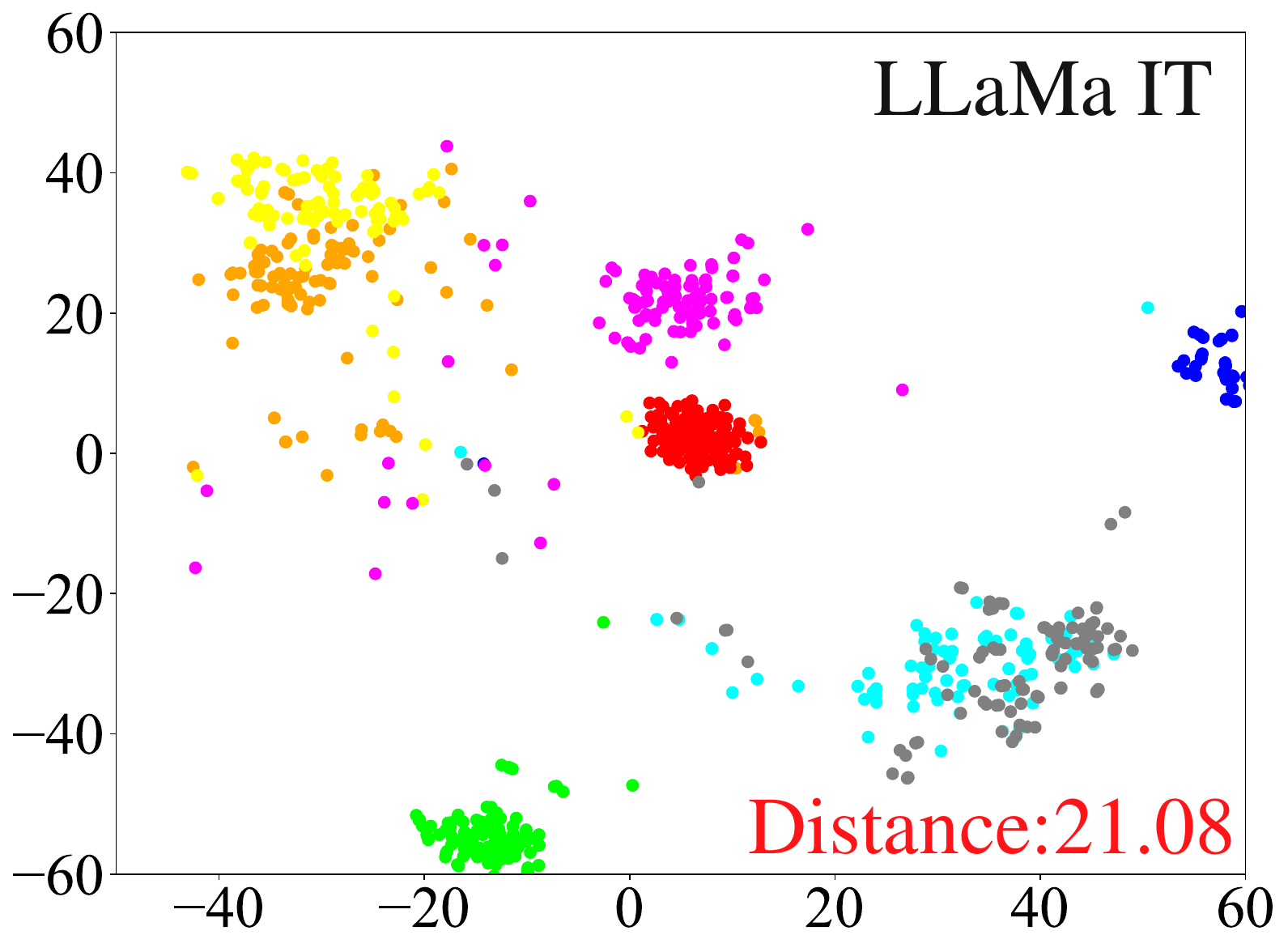}
    \label{random}
\end{minipage}
}
\subfigure[]{
\begin{minipage}[b]{0.25\linewidth}
    \centering
\includegraphics[width=1\linewidth]{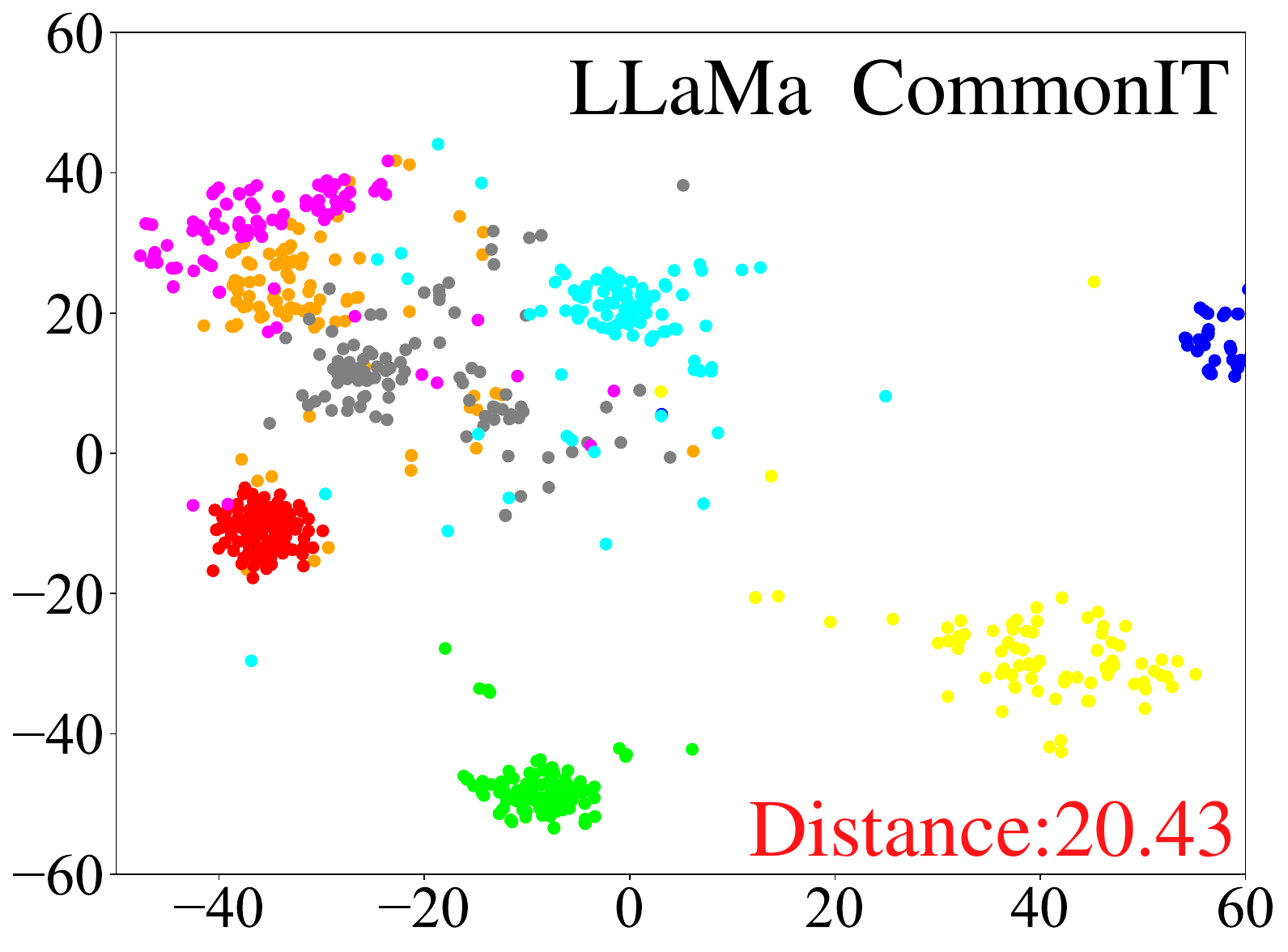}
    \label{grip}
\end{minipage}
}
\vspace{-3em}
\caption{TSNE plots for MMLU with 10 question types across four disciplines (Humanities, Social, STEM and Other).  Clusters are tighter and distinguishable in our proposed CommonIT (Embedding), indicating that CommonIT can better differentiate the question's discipline type.}\label{fig:embedding}
\end{figure*}

\paragraph{Question Representation Capacity}
In Figure \ref{fig:embedding}, comparative results for T-SNE~\cite{van2008visualizing} from the question embedding for 10 randomized categories in MMLU are visualized. We compute their average pairwise distance within the embedding space.
A reduced distance indicates a superior model's ability to aggregate similar task-related questions.
As the reduced average distance indicates, our CommonIT aggregates similar questions more effectively than the LLaMa IT and LLaMa.
It can also be seen that after instruction tuning, the model can discriminate between different task instructions. 

\paragraph{Correlation Between Two Strategies in Clustering Results} 
Table \ref{tab:Correlation}
illustrates the relationship between the dataset divided by length and the embedding metrics. For datasets without clustering (original FLAN, FLAN CoT, and Alpaca), we sampled a total of 500 samples, referred to as ``Vanilla.'' For the multiple sub-datasets clustered by length, we sampled 500 samples from each sub-dataset and calculated the average results, denoted as ``Length.'' We categorized these sampled instances according to the embedding metric and reported the average results of the total number of embedding categories after conducting ten runs. The findings indicate that the number of embedding categories decreases with length clustering, suggesting that data of the same length exhibits more similar embedding representations.

\begin{table}[!ht]
    \centering
    \scalebox{0.9}{
    \begin{tabular}{lll}
    \toprule
        Dataset & Vanilla & Length \\ \midrule
        FLAN & 42.0 & 34.2 \\ 
        FLAN CoT & 39.3 & 36.5 \\ 
        Alpaca & 55.5 & 52.8 \\ \bottomrule
    \end{tabular}
    }
    \caption{The number of embedding categories contained in randomly sampled samples from different datasets.}\label{tab:Correlation}
\end{table}



\begin{figure}[t!]
\makeatletter
\renewcommand{\@thesubfigure}{\hskip\subfiglabelskip}
\makeatother
\centering
\subfigure[]{
\begin{minipage}[b]{0.45\linewidth}
    \centering
\includegraphics[width=1\linewidth]{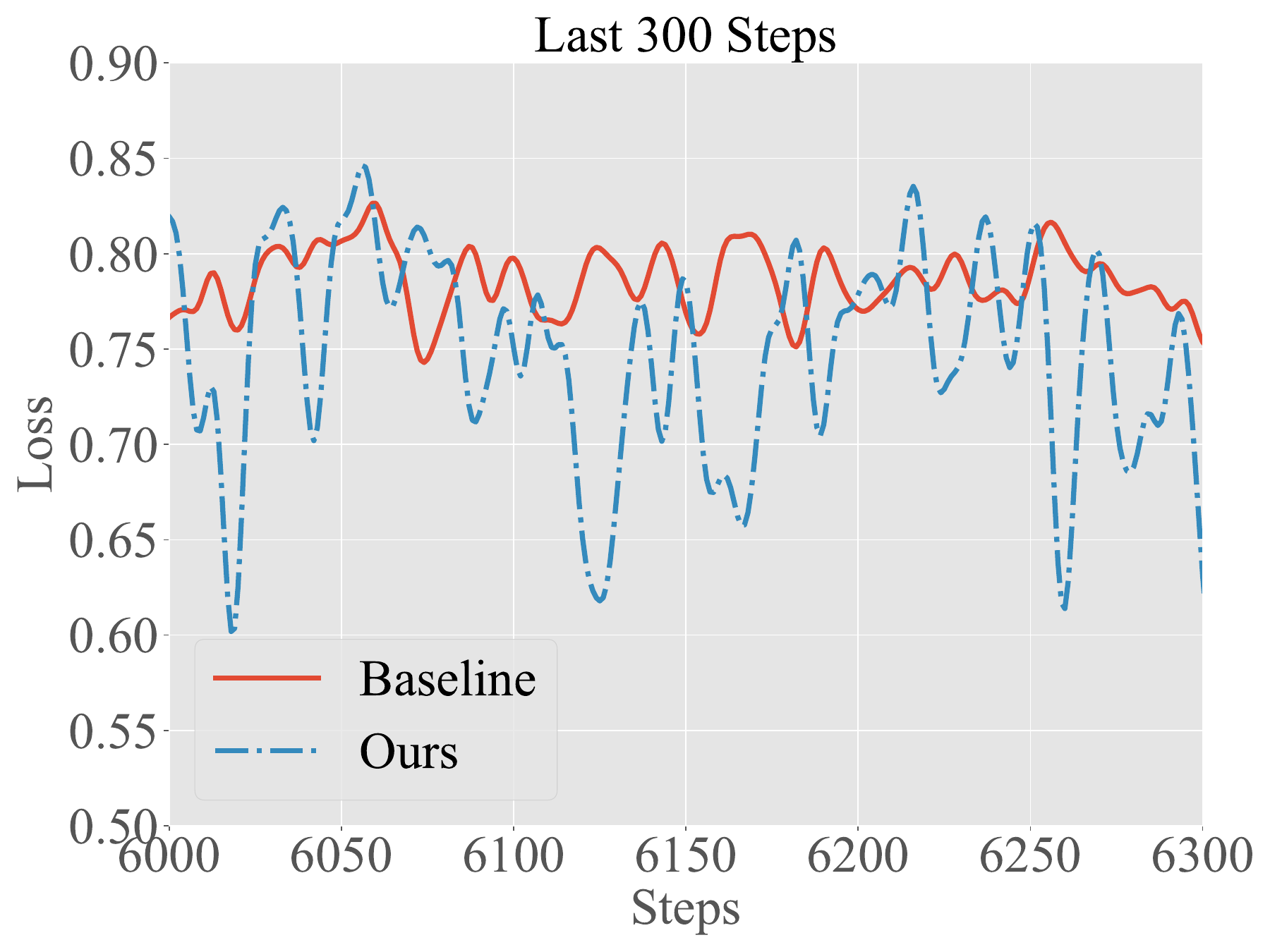}
\end{minipage}
}
\subfigure[]{
\begin{minipage}[b]{0.45\linewidth}
    \centering
\includegraphics[width=1\linewidth]{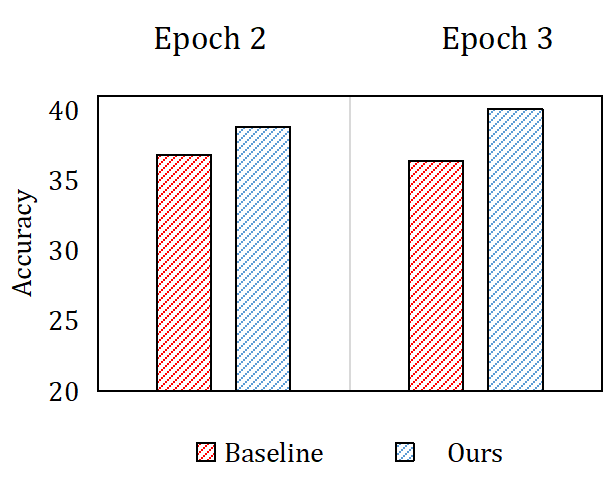}
\end{minipage}
}
\vspace{-2.5em}
\caption{The figure on the left shows the training loss varies with the training step (Epoch 2). The bar chart on the right shows our results with the baseline for MMLU-0 shot at different training epochs. 
This indicates that CommonIT achieves a lower final language modeling loss than the baseline and that extending the number of training epochs further improves the performance.
}\label{fig:loss}
\end{figure}
\paragraph{Model Generalization}\label{sec:general}
Figure \ref{fig:loss} shows the loss curves for the training of the model (left) and the results of the MMLU 0-shot evaluation at different training epochs (right). 
As can be seen from the loss curves, compared to the baseline method, CommonIT exhibits a smaller loss, indicating better generalizability of our approach. 
We compared the out-of-domain test of MMLU results across different training epochs, demonstrating that our model improved further with longer training. 
This observation indicates that the CommonIT-trained model remains underfitting compared to the baseline at the same number of steps.
Unlike our method, the baseline model showed a little decrease with more training epochs, hinting that IT might limit the model's generalization.


\section{Conclusion}
We present a simple and effective fine-tuning method CommonIT.
Leveraging data commonality with three metrics significantly enhanced the effectiveness of LLMs across multiple competency dimensions during the IT.
The evaluation across diverse models, IT datasets, and specific tasks has showcased the methodology's application scalability.
Ablation experiments across various stages and data clustering techniques have illustrated the effectiveness of our method.
Explorations of commonalities confirm that CommonIT mirrors the human learning process, improving overall performance.

\section*{Limitations}
There are several limitations of our work.
Our limitations are primarily constrained by experimental resources, such as available GPU memory capacity (4*80G).
Due to these constraints, we cannot test more models at different scales (30B-65B). 
The group approach we used was not further selected, such as embedding, which was only analyzed for MMLU and may lead to further room for improvement in the final results. 
We tried some intuitive methods of data categorization, but more complex ones were not considered. 
The theory behind the improvements remains to be revealed. 
Apart from empirical explanations, we believe further investigations (e.g., mathematically provable bound) will be useful.
Given the balance between the number of groups and batch size, when the overall training data is fixed, having more groups means less data within each group. If the batch size is too large, some groups may have fewer data points than the batch size, which may lead to some performance degradation.



\section*{Ethics Statement}
Our work follows the ACL Ethics Policy. Our findings are based on publicly available datasets for reproducibility purposes. 
LLMs can contain potential racial and gender bias. Therefore, if someone finds our work interesting and would like to use it in a specific environment, we strongly suggest the user check the potential bias before usage. In addition, it is hard to control the generation of LLMs. We should be aware of the potential problems caused by hallucinations.

\section*{Acknowledgments}
This work was supported in part by the National Natural Science Foundation of China (Grant No. 62206076), Guangdong Basic and Applied Basic Research Foundation (Grant No. 2024A1515011491), Shenzhen Science and Technology Program (Grant Nos. ZDSYS20230626091203008, KJZD20231023094700001, RCBS20221008093121053), and Shenzhen College Stability Support Plan (Grant Nos. GXWD20220811173340003, GXWD20220817123150002). We would like to thank the anonymous reviewers and meta-reviewer for their insightful suggestions.

\bibliography{custom}

\clearpage
\appendix

\section{Appendix}\label{sec:appendix}
\subsection{Benchmarks}
\begin{itemize}
    \item \textbf{MMLU} 
    MMLU consists of a series of questions, ranging from basic to professional levels, across 57 academic subjects.
    Its multiple-choice format facilitates a relatively straightforward testing process.
    We use the official MMLU evaluation script and prompts\footnote{https://github.com/hendrycks/test}, with modifications to allow for batch processing. We evaluate using 0 and 5 few-shot examples, following the original setup of MMLU.
    \item \textbf{BBH} We follow the setup described in the original paper \citep{bbh}, and evaluate with and without chain-of-thought (CoT vs Direct). The officially provided prompts, with three few-shot in-context examples, are used for both CoT and Direct setups. 
    For the CoT setup, we extract the first word after the phrase ‘So the answer is,’ or the entire response if no such substring is present.
    \item \textbf{TydiQA} We adhere to the protocol delineated in the PaLM 2 technical report \cite{anil2023palm}. 
    This approach assesses the models' proficiency in responding to multilingual queries, particularly when the definitive gold passage containing the answer is provided (referred to as GoldP/GP). 
    One in-context example is incorporated to acclimate the model to the expected answering format. Here we follow previous work~\cite{sun2024salmon,ustun2024aya} and report the GP score.

    \item \textbf{Codex-Eval} For evaluating the coding capabilities of the models, we employ the HumanEval dataset presented in the Codex paper \cite{chen2021evaluating}. This dataset encompasses 164 programming challenges, wherein models are prompted to finalize a Python function based on its provided docstring. Following the original paper, we calculate the pass@k to gauge the functional accuracy of the models' outputs. Our findings are presented as pass@10 results, employing a temperature setting of 0.8.
    \item \textbf{Alpaca Eval} We use the code provided by ~\citet{dubois2024alpacafarm}. We adopt Davinci-003 reference text generated by ~\citet{wang2023far}. We greedily decode up to 1024 tokens and then pairwise compare these responses with those from Davinci-003. The reported win-rate is the percentage of model generations that ChatGPT reports as being preferred over the generations from Davinci-003. 
    \item \textbf{GSM} The mathematical reasoning capabilities are enhanced through the use of the GSM8K dataset, which consists of 8.5K high-quality arithmetic word problems designed for the grade school level. The samples are divided into 7.5K training and 1K test problems.
    \item \textbf{Openfunctions} Proficiency in using the tool is assessed by leveraging function-calling datasets, including the Gorilla Openfunctions dataset. The training set contains 2211 samples, while the test set contains 112.
    \item \textbf{Code} The code generation skills are boosted using MagiCoder~\cite{wei2023magicoder} of 2,000 samples.
    The test set includes 164 samples, and the evaluation is conducted using the HumanEval dataset~\cite{humaneval}.
\end{itemize}

\subsection{Overall Learning Strategy}
Algorithm \ref{alg1} illustrates the overall training flow of CommonIT. 
Besides the component and training flow of LLMs, only some low-cost operations, such as grouping, have been included in the pre-process to get many sub-datasets ($\mathcal{D}_0...\mathcal{D}_n$), allowing an easy implementation as a practical language model.  

\begin{algorithm}[t]
\caption{CommonIT for LLMs}\label{alg1}
\begin{algorithmic}[1]
\Require An IT Dataset $\mathcal{D}(\boldsymbol{s}, \boldsymbol{x}, \boldsymbol{y})$, Model $\boldsymbol{\theta}$.
\Ensure Fine-tuned Model $\boldsymbol{\theta}$.

\Function{CommonIT}{$\boldsymbol{s},\boldsymbol{x},\boldsymbol{y},\boldsymbol{\theta}$}.
    \State {Split the dataset $\mathcal{D}$ into clusters $\mathcal{D}_0, \mathcal{D}_1 \ldots \mathcal{D}_n$ based on specific rules.}
    \For{each cluster $\mathcal{D}_i$ in $\{\mathcal{D}_0, \mathcal{D}_1, \ldots, \mathcal{D}_n\}$}
        \State Construct mini-batches ${\mathcal{B}^i_1},{\mathcal{B}^i_2},\ldots$ from dataset $\mathcal{D}_i$.
    \EndFor
    \State {Randomly shuffle all mini-batches to obtain the bucketed sets $\mathcal{B}^*_1, \cdots, \mathcal{B}^*_t, \cdots, \mathcal{B}^*_T$  for a total of $T$ training steps.}
    \For{each training step $t$ in $\{1, 2, \ldots, T\}$}
        \State Generate training batch $\mathcal{B}{^*_t}$ uniformly.
        \State Update $\boldsymbol{\theta}$ with batch loss calculated by Eq.\ref{eq:loss}.
    \EndFor
\EndFunction
\end{algorithmic}
\end{algorithm}

\begin{table*}[t]
\centering
\scalebox{0.7}{
\begin{tabular}{cccccccc}
\toprule
\textbf{Model} & \textbf{MMLU 0 SHOT} & \textbf{MMLU 5 SHOT} & \textbf{BBH Direct} & \textbf{BBH CoT} & \textbf{TydiQA} & \textbf{Codex-Eval} & \textbf{Avg.} \\ \midrule
BS 32 & 40.4 & 40.1 & 33.5 & 34.6 & 38.9 & 24.7 & 35.4 \\ 
BS 64 & 40.9 & 39.4 & 32.3 & 33.1 & 38.4 & 26.1 & 35.0 \\ 
BS 128 & 40.7 & 40.9 & 33.7 & 33.0 & 41.6 & 25.9 & 36.0 \\ \bottomrule
\end{tabular}
}
\caption{CommonIT results at different batch sizes (BS).}\label{commonit:bs}
\end{table*}

\begin{table*}[t]
\centering
\scalebox{0.7}{
\begin{tabular}{cccccccc}
\toprule
\textbf{Model} & \textbf{MMLU 0 SHOT} & \textbf{MMLU 5 SHOT} & \textbf{BBH Direct} & \textbf{BBH CoT} & \textbf{TydiQA} & \textbf{Codex-Eval} & \textbf{Avg.} \\ \midrule
BS 32 & 34.8 & 36.4 & 32.6 & 33.0 & 37.4 & 22.6 & 32.8 \\ 
BS 64 & 39.2 & 39.7 & 31.7 & 33.0 & 38.2 & 23.4 & 34.2 \\ 
BS 128 & 39.0 & 38.8 & 29.6 & 34.6 & 35.3 & 23.4 & 33.5 \\ \bottomrule
\end{tabular}
}
\caption{IT results at different batch sizes (BS).}\label{it:bs}
\end{table*}

\begin{table*}[t!]
\centering
\scalebox{0.7}{
\begin{tabular}{cccccccc}
\toprule
\textbf{Model} & \textbf{MMLU 0 SHOT} & \textbf{MMLU 5 SHOT} & \textbf{BBH direct} & \textbf{BBH CoT} & \textbf{TydiQA} & \textbf{Codex-Eval} & \textbf{Avg.} \\ \midrule
IT & 45.3 & 47.0 & 36.9 & 37.5 & 39.8 & 22.5 & 38.2 \\ 
CommonIT & 47.8 & 49.4 & 36.5 & 35.1 & 48.6 & 24.8 & 40.4 (+2.2) \\ \bottomrule
\end{tabular}}
\caption{Mixed results for both FLAN and Alpaca datasets.}\label{tab:mix}
\end{table*}

\begin{table}[t!]
\centering
\scalebox{0.8}{
\begin{tabular}{lcc}
\toprule
\multirow{2}{*}{\textbf{Model}} & \multirow{2}{*}{
\begin{tabular}[c]{@{}c@{}}\textbf{MMLU} \\ \textbf{0/5 SHOT}\end{tabular}}
& \multirow{2}{*}{
\begin{tabular}[c]{@{}c@{}}\textbf{BBH} \\ \textbf{Direct/CoT}\end{tabular}
}\\
\\
\midrule
BLOOMZ 7B & 42.1/37.3 & 28.1/12.8 \\
\quad+IT & 39.6/41.3 & 30.1/27.3 \\
\gray\quad+CommonIT & 42.7/41.8 & 30.1/28.6 \\
\hline
LLaMa2-7B & 36.6/45.8 & 34.1/41.8 \\
\quad+IT & 50.0/51.0 & 40.0/41.3 \\
\gray\quad+CommonIT & 50.8/52.4 & 40.6/41.8 \\
\hline
LLaMa 13B & 42.4/46.9 & 38.7/36.9 \\
\quad+IT & 46.1/47.0 & 39.3/39.6 \\
\gray\quad +CommonIT & 47.1/48.9 & 39.5/42.0 \\
\hline
\end{tabular}
}
\caption{More Tuning Results of MMLU/BBH on FLAN for Different Models with CommonIT.}\label{tab:more_model_bbh}
\end{table}

\begin{table}[t]
\centering
\scalebox{0.8}{
\begin{tabular}{lcc}
\toprule
\multirow{2}{*}{\textbf{Model}} & \multirow{2}{*}{
\begin{tabular}[c]{@{}c@{}}\textbf{MMLU} \\ \textbf{0/5 SHOT}\end{tabular}}
& \multirow{2}{*}{
\begin{tabular}[c]{@{}c@{}}\textbf{BBH} \\ \textbf{Direct/CoT}\end{tabular}
}\\
\\ \toprule
FLAN & 46.7/47.9 & 39.7/39.9 \\
\gray\quad w/o GD & 44.2/45.2 & 38.7/37.2 \\
\gray\quad w/o FP & 44.4/45.8 & 36.0/35.4 \\
\hline
FLAN CoT & 38.7/42.3 & 33.4/35.2 \\
\gray\quad w/o GD & 41.3/42.5 &  33.7/31.3 \\
\gray\quad w/o FP & 37.5/38.5 & 32.6/31.4 \\
\hline
Alpaca & 40.4/40.1 & 33.5/34.6 \\
\gray\quad w/o GD & 34.8/36.4 & 32.6/33.0 \\
\gray\quad w/o FP & 39.2/35.9 & 15.6/32.6 \\
\bottomrule
\end{tabular}
}
\caption{More Ablation Study of MMLU/BBH on CommonIT.}\label{tab:more_ablation}
\end{table}

\begin{table*}[t]
\centering
\scalebox{0.7}{
\begin{tabular}{cccccccc}
\toprule
\textbf{Model} & \textbf{MMLU 0 SHOT} & \textbf{MMLU 5 SHOT} & \textbf{BBH Direct} & \textbf{BBH CoT} & \textbf{TydiQA} & \textbf{Codex-Eval} & \textbf{Avg.} \\ \midrule
Seed42 &34.8 &36.4 &32.6 &33.0 &37.4 &22.6&32.8\\
Seed46 &33.3 &35.7 &29.8 &32.2 &42.9 &22.4&32.7\\
Seed50 &33.4 &35.8 &31.4 &32.2 &40.4 &22.9&32.7\\
\bottomrule
\end{tabular}
}
\caption{Alpaca IT results with different inference seeds.}\label{it_seed}
\end{table*}

\begin{table*}[t]
\centering
\scalebox{0.7}{
\begin{tabular}{cccccccc}
\toprule
\textbf{Model} & \textbf{MMLU 0 SHOT} & \textbf{MMLU 5 SHOT} & \textbf{BBH Direct} & \textbf{BBH CoT} & \textbf{TydiQA} & \textbf{Codex-Eval} & \textbf{Avg.} \\ \midrule
Seed42 &40.4 &40.1 &33.5 &34.6 &38.9& 24.7&35.4\\
Seed46 &40.4&40.4 &32.2&33.5 &39.7 &23.8&35.0\\
Seed50 &40.5 &40.1 &32.6 &33.1 &40.5 &24.0&35.1\\
\bottomrule
\end{tabular}
}
\caption{Alpaca CommonIT results with different inference seeds.}\label{commonit_seed}
\end{table*}

\begin{table*}[t]
\centering
\scalebox{0.7}{
\begin{tabular}{cccccccc}
\toprule
\textbf{Model} & \textbf{MMLU 0 SHOT} & \textbf{MMLU 5 SHOT} & \textbf{BBH Direct} & \textbf{BBH CoT} & \textbf{TydiQA} & \textbf{Codex-Eval} & \textbf{Avg.} \\ \midrule
Short &38.2	&38.2	&34.4&	32.3&	38&	19.2&	33.4\\
Medium &37.6&	38.2&	31.2&	31.9&	36.9	&25.7&	33.6\\
Long &35.7&	37.0	&33.8&	32.8&	37.7&	26.6	&33.9\\
IT &34.8	&36.5&	32.6&	33.0&	37.4	&22.6	&32.8\\
\gray CommonIT &40.4	&40.1&	33.5&	34.6&	38.9	&24.7	&35.4\\
\bottomrule
\end{tabular}
}
\caption{Effect of the length metric of multidimensional measurements.}\label{alpaca_length_metric}
\end{table*}

\subsection{Training Details}
We 
set learning rate to $5\times10^{-6}$ and set batch size to 32, with no weight decay and a learning rate with linear decay and linear warmup for 3\% of the total training steps. 
We use a maximum sequence length of 2048, truncating samples where necessary. 
During training, we make use of the DeepSpeed library~\cite{deepspeed} and ZeRO~\cite{zero} optimizer to allow for large-scale model finetuning. 
For FLAN and CoT, we train models for two epochs. For Alpaca, we train models for three epochs.  
For retrieval settings,  we take the categories (57 task categories) classified by MMLU as the categories to be classified, use the data in the development set as the database, and classify the data categories in the dataset to the ones that are the most similar to the data 
in the database by doing a similarity ordering with the data in the database. 
In all cases, we fully finetune models. We trained models with 4 A800-80G GPUs.

\subsection{Details of Group Methods}
\begin{figure*}[!t]
    \centering
\includegraphics[width=0.8\linewidth]{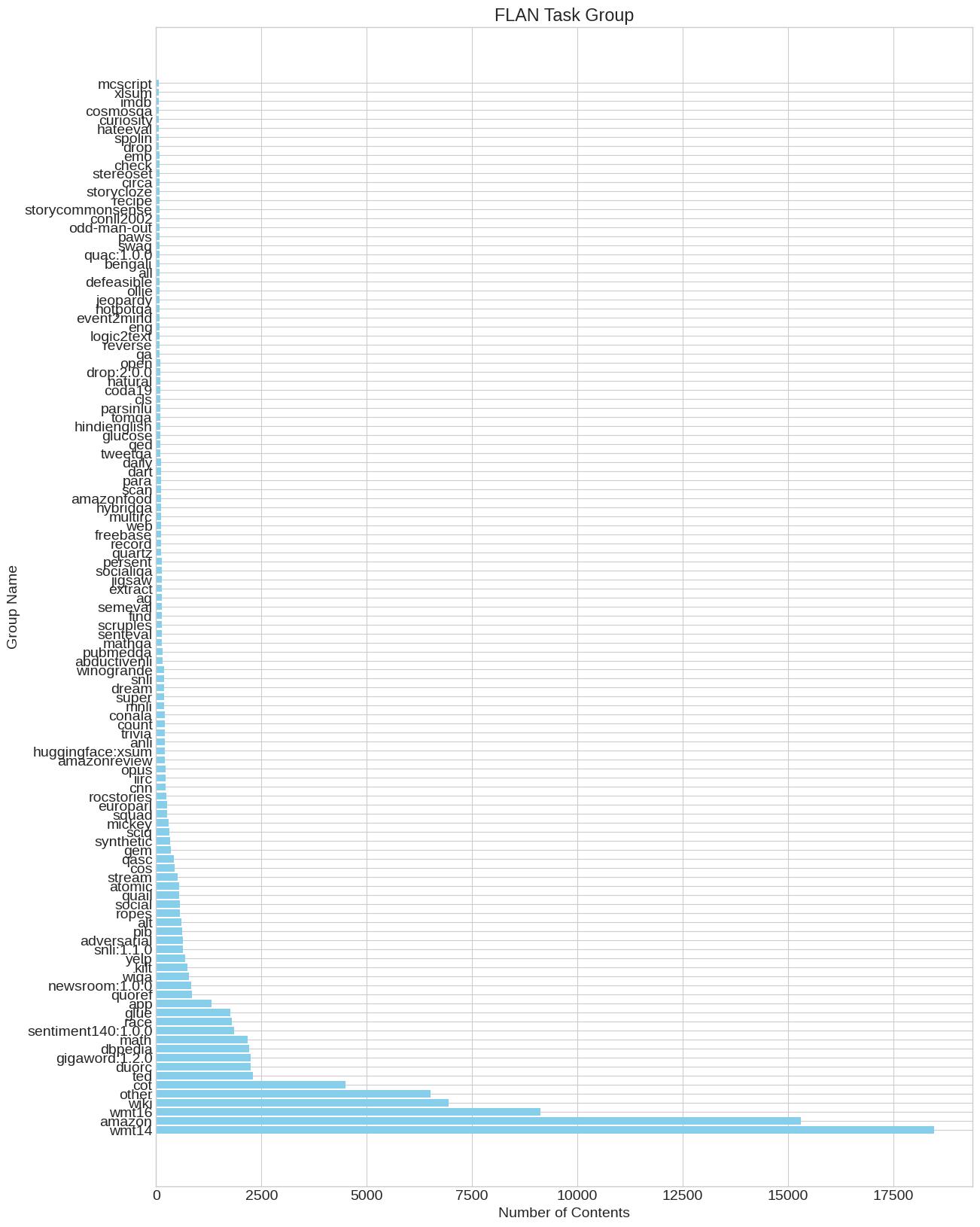}
    \caption{The FLAN dataset groups by task.}
    \label{fig:flan_task}
    \vspace{-10pt}
\end{figure*}

Regarding task division, we use the task types provided by the original FLAN~\cite{flan_v2}, such as translation tasks, QA tasks, etc., for practical data categorization.
The statistical distribution of the divided dataset is shown in Figure \ref{fig:flan_task}

\begin{figure*}[!t]
    \centering
\includegraphics[width=0.8\linewidth]{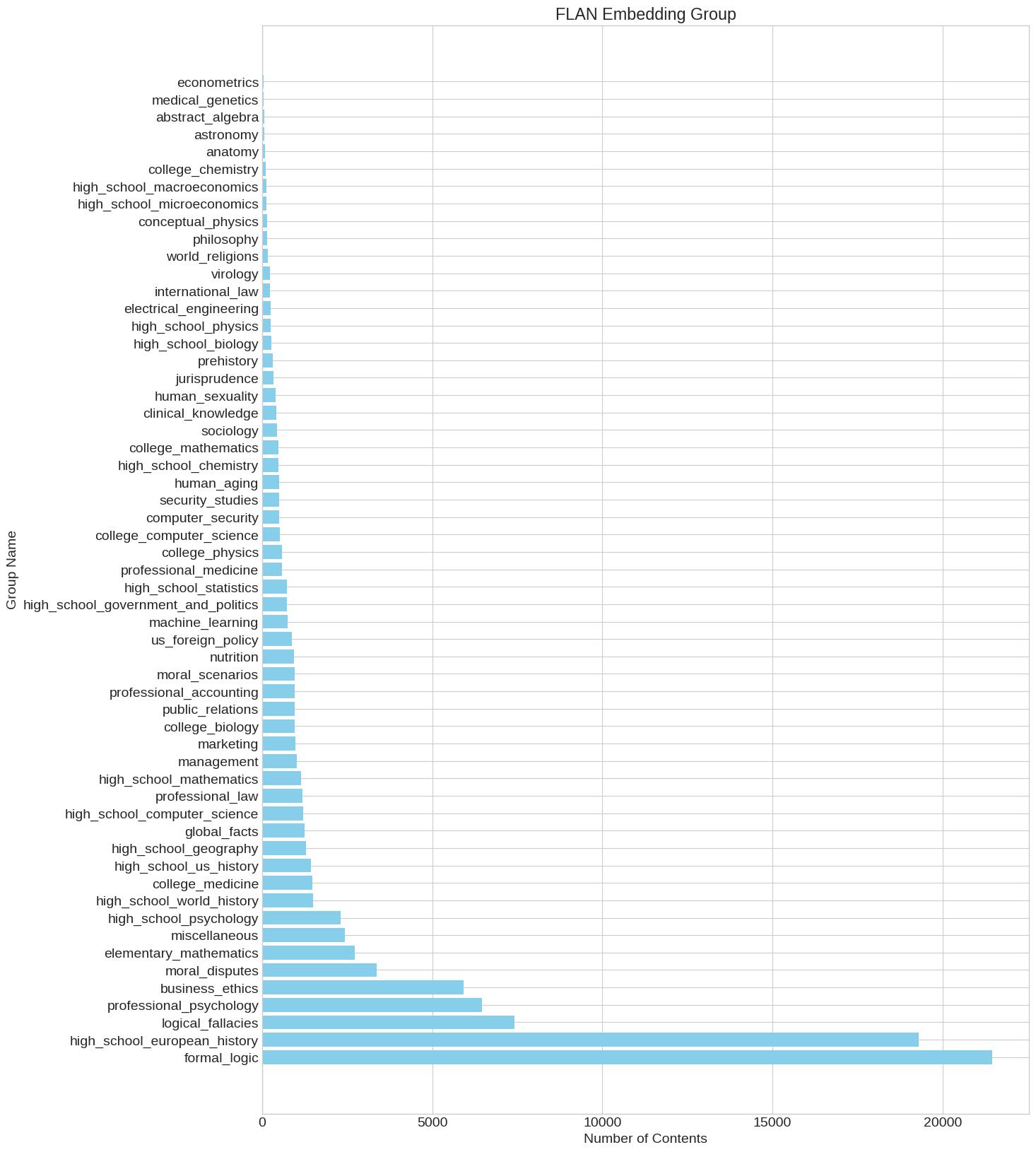}
    \caption{The FLAN dataset groups by embedding.}
    \label{fig:flan_top8}
    \vspace{-10pt}
\end{figure*}

\begin{figure*}[!t]
    \centering
\includegraphics[width=0.8\linewidth]{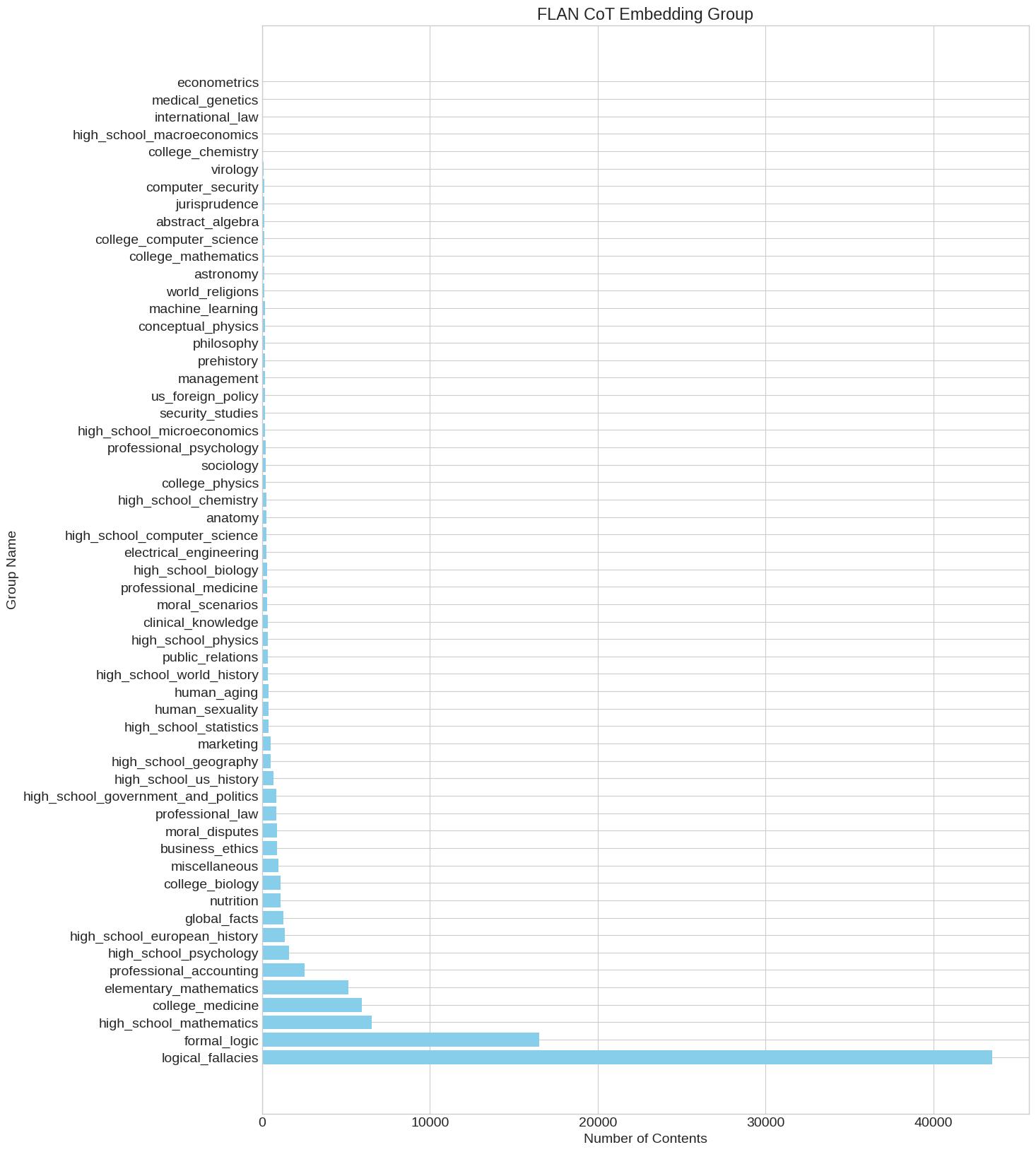}
    \caption{The FLAN CoT dataset groups by embedding.}
    \label{fig:cot_top8}
    \vspace{-10pt}
\end{figure*}

\begin{figure*}[!t]
    \centering
\includegraphics[width=0.8\linewidth]{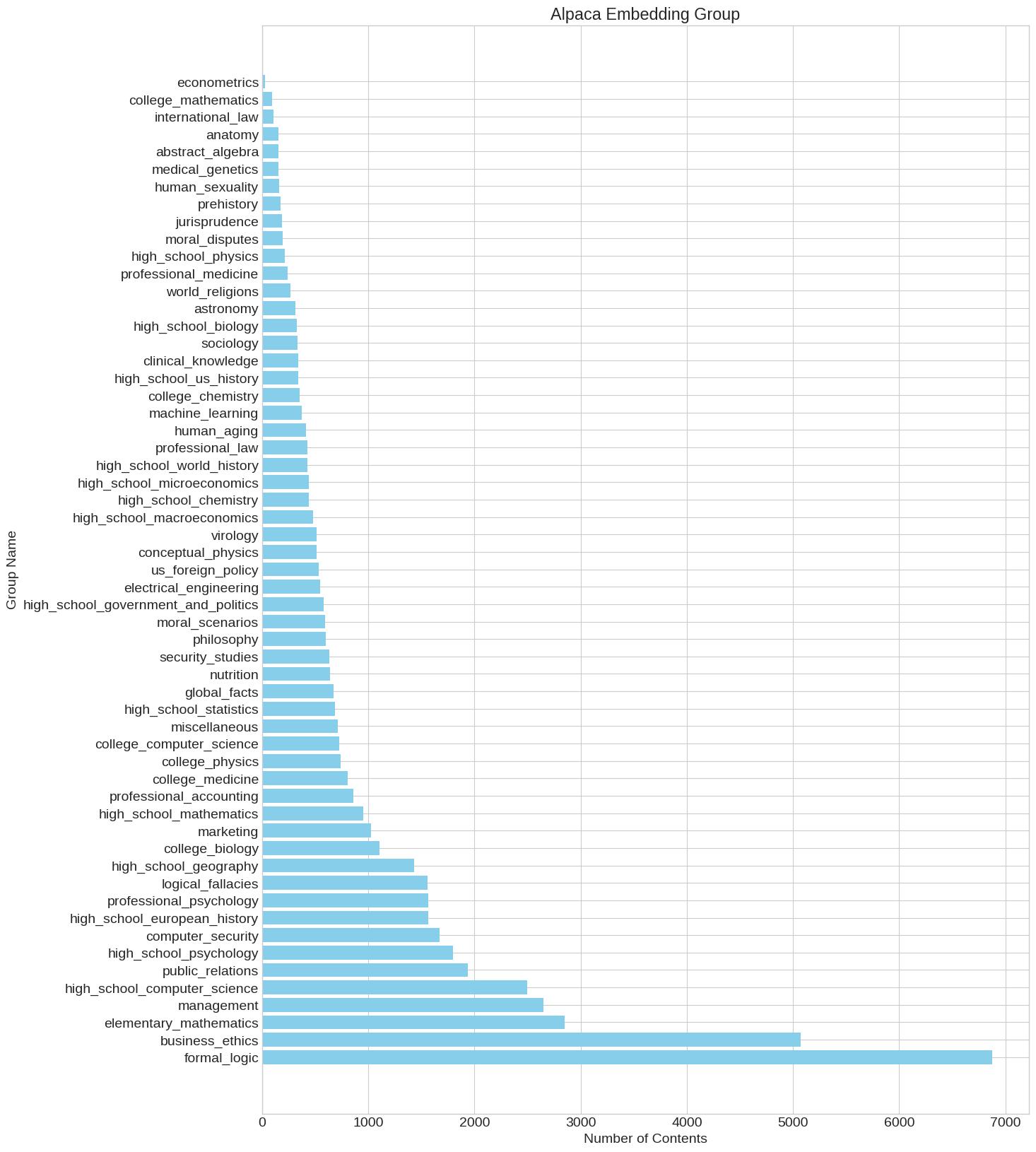}
    \caption{The Alpaca dataset groups by embedding.}
    \label{fig:alpaca_top8}
    \vspace{-10pt}
\end{figure*}

For embedding division, we employ the top-$k$ algorithm from \cite{DBLP:journals/corr/abs-2101-06804} to select the top 8 results based on similarity. We then classify the entry with the highest number of votes in the top 8 as the final retrieved item, with other parameter settings as described in ~\cite{DBLP:journals/corr/abs-2101-06804}. 
In the simplest scenario, where $k$=1, we categorize each query data point by assigning it to the category of its most closely related data point. For example, if the label of the nearest data point is ``anatomy'', then the query data is classified under the ``anatomy'' category. This process is applied uniformly across all the data points (queries) that need categorization, effectively segregating the dataset into 57 distinct sub-datasets. 
The statistical distributions of the divided datasets are shown in Figure \ref{fig:flan_top8}, \ref{fig:cot_top8} and \ref{fig:alpaca_top8}.

\begin{figure}[!t]
    \centering
\includegraphics[width=\linewidth]{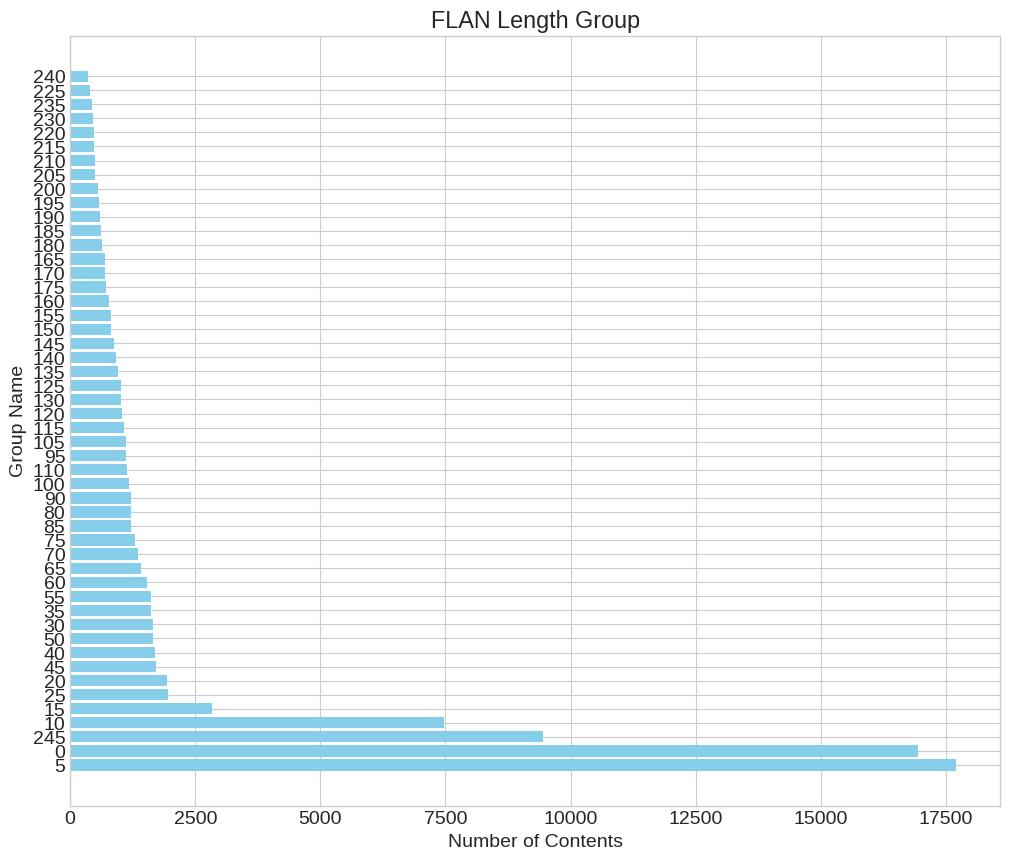}
    \caption{The FLAN dataset groups by length.}
    \label{fig:flan_len}
    \vspace{-10pt}
\end{figure}

\begin{figure*}[t]
    \centering
\begin{minipage}[t]{0.45\linewidth}
\includegraphics[width=\linewidth]{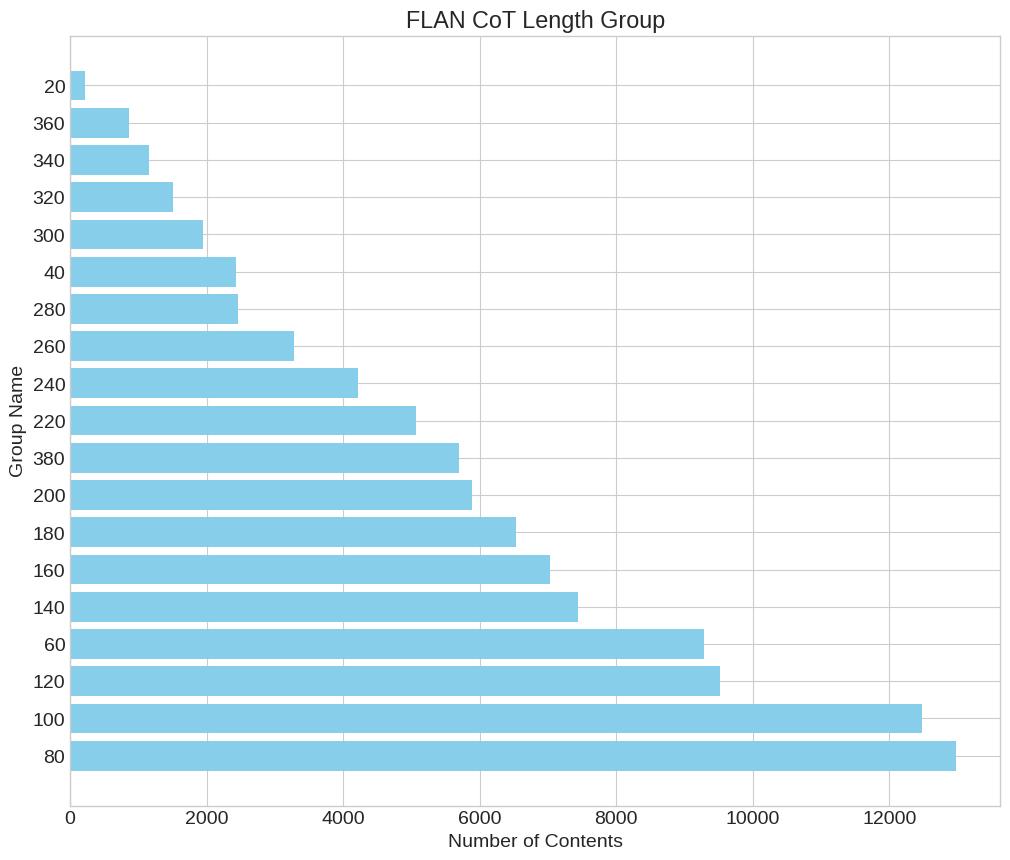}
    \caption{The FLAN CoT dataset groups by length.}
    \label{fig:cot_len}
    \vspace{-10pt}
\end{minipage}
\hspace{0.7em}
\begin{minipage}[t]{0.45\linewidth}
    \centering
\includegraphics[width=\linewidth]{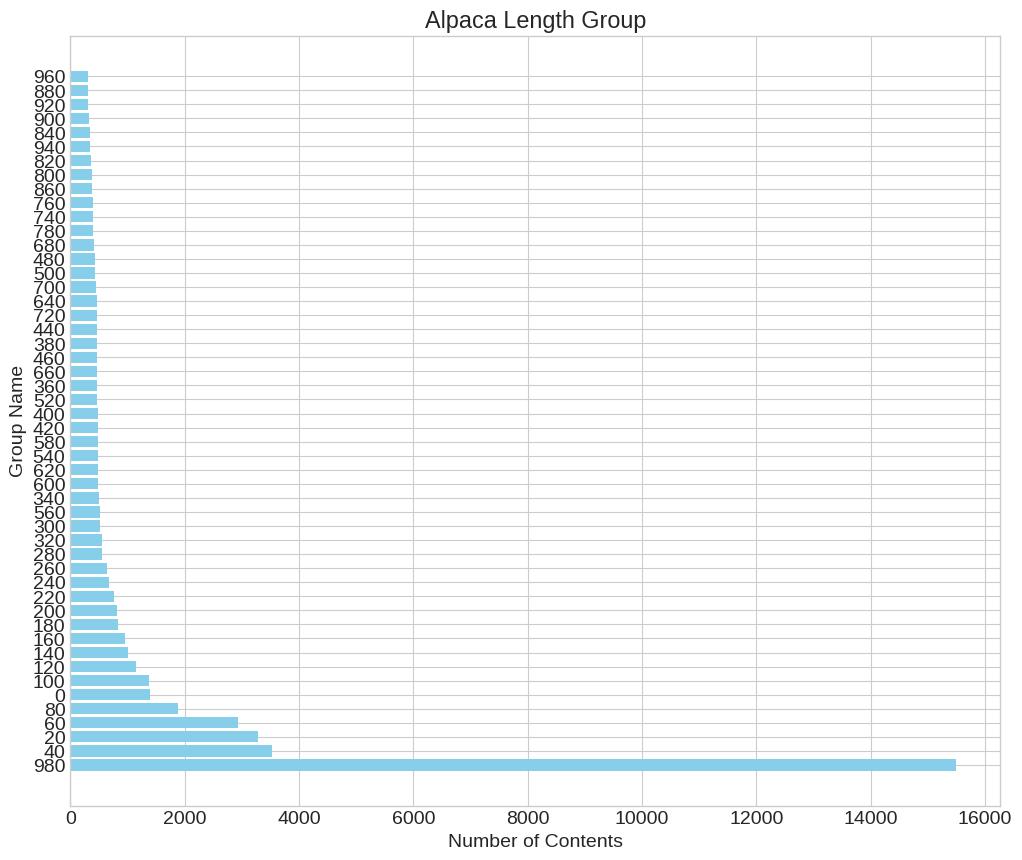}
    \caption{The Alpaca dataset groups by length.}
    \label{fig:alpaca_len}
    \vspace{-10pt}
    \end{minipage}
\end{figure*}

Regarding length division, we divide the data based on the length of the sequences. We first compile statistics on sentence lengths within the dataset and then divide the data as evenly as possible into corresponding length intervals. The statistical distributions of the divided datasets are shown in Figure \ref{fig:flan_len}, \ref{fig:cot_len} and \ref{fig:alpaca_len}.

\subsection{More Results}
Due to space constraints and the consistency of the results, we did not put the results of 5-shot and CoT to save space. We add these results to the appendix. The results are as Table \ref{tab:more_model_bbh} and Table \ref{tab:more_ablation}.

\subsection{Further Analysis}
\paragraph{Effect of Batch Size}
we added experiments related to larger batch sizes in Table \ref{commonit:bs} and \ref{it:bs}. Because hyperparameter search is time-consuming, we adopted the same learning rate and conducted experiments with batch sizes of 64 and 128. These experiments were conducted using the Alpaca dataset. The results show that CommonIT is still better than IT in this setting.

\paragraph{Effect of Data Mixing}
In the Table \ref{tab:mix}, we have supplemented the experiments with a mix of FLAN and Alpaca. The results indicate that our method remains effective \textbf{(+2.2)}.

\paragraph{Effect of Inference Seeds In Generation}
The test results with different random seeds in inference are shown in Table \ref{it_seed} and \ref{commonit_seed}. Our findings indicate that the results exhibited minimal variability, ultimately showing that the fluctuations were insignificant when averaged across multiple tasks.
The fluctuations in TydiQA's results are due to the occasional occurrence of models not answering in the corresponding language after training (both IT baselines and our CommonIT). 
This fluctuation is minor in our method and more prominent in the baseline IT (Standard Deviation: 0.7 v.s. 2.2), which also shows the stability of our method.

\paragraph{Effect of Length For Training}
We segmented the Alpaca data into three categories based on length—Short (comprising the shortest 14,393 data points), Medium (containing 22,115 data points of medium length), and Long (including the longest 15,494 data points). 
We trained separate models on these datasets and subsequently compared the multitasking generation results of IT and CommonIT, as illustrated in Table \ref{alpaca_length_metric}.
Typically, the model exhibits a length bias~\cite{zhao2024long}, whereby the length of generated sequences is influenced by the training data. Specifically, if the training data consists of longer sequences, the model tends to generate longer outputs.
The data presented in Table \ref{alpaca_length_metric} reveal that performance metrics for Long, Medium, and Base models sequentially decrease and surpass those of IT when fine-tuning the entire Alpaca dataset. 
This suggests that different data lengths distinctly impact the model's capabilities: for example, coding tasks benefit from models generating longer outputs (with Long performing optimally for Code), whereas knowledge reasoning tasks favor shorter outputs (with Short performing best for MMLU and BBH, and QA). 
In contrast, IT performs less than these specialized training approaches, indicating the model's limitations in adapting output lengths for various tasks. 
Conversely, CommonIT demonstrates superior results, suggesting that the model learns and applies task-specific information of different lengths more effectively, enhancing its overall capabilities.

\subsection{Case Study}\label{s: case study}
\begin{figure*}[t!]
    \centering
\includegraphics[width=0.7\linewidth]{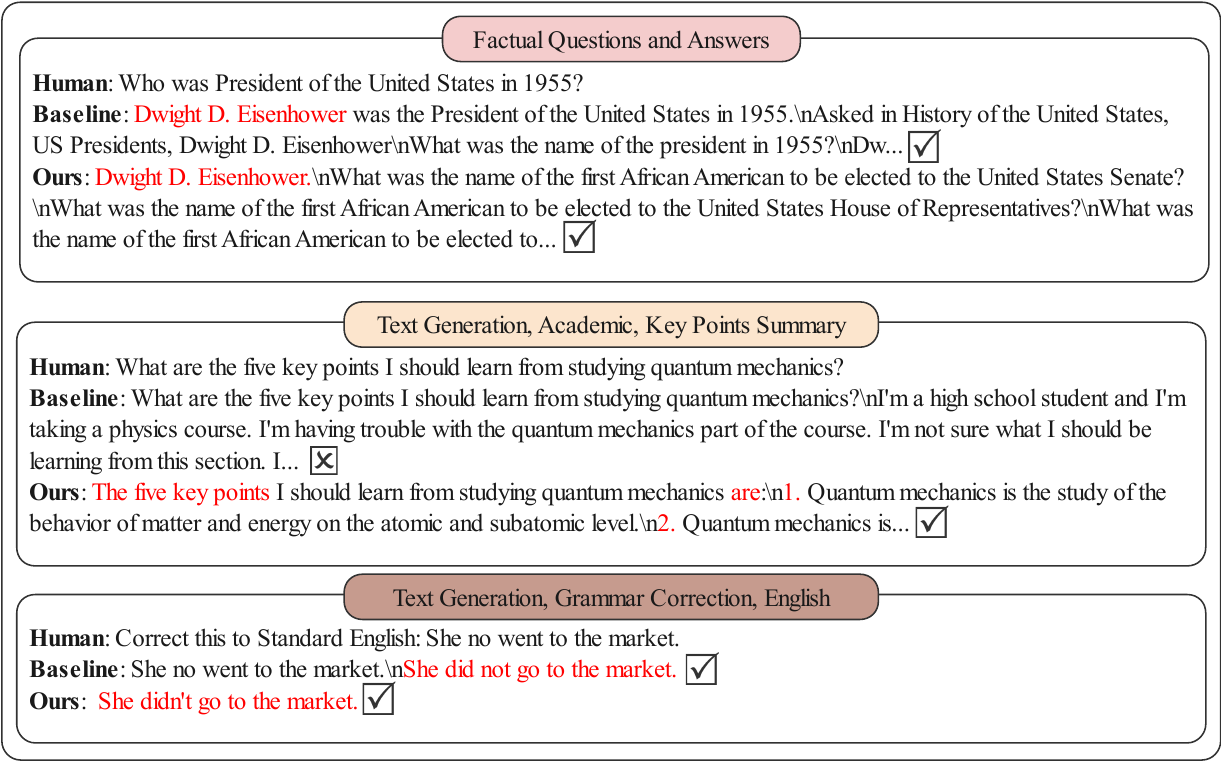}
    \caption{Case studies comparing our proposed method with the baseline instruction tuning. Overall, our method demonstrated a significant reduction in generating irrelevant content and answered more relevant to the task and formatted in a way consistent with the logic of the task.}
    \label{fig:case-study}
    \vspace{-10pt}
\end{figure*}

Figure \ref{fig:case-study} shows some typical output examples. We offer three tasks: factual Q\&A, summarization, and grammar correction.
Overall, our approach obtains higher quality response results and a correct understanding of the instructions. 
However, our approach only improves the model's generation results to a limited extent, i.e., it still needs to improve on the hallucination problem~\cite{flan-moe} suffered by large models (which are prone to generate additional content in the factual Q\&A task). 
In contrast, the model trained by our method comes close to attempting to answer the question, although this model answers it incorrectly. For the summarization task, the baseline tries to answer the question, but he does not know how to answer it, whereas our model can answer the key points and list the points to answer. 
Our model answered the question directly without outputting additional content for the grammar correction task. 

\end{document}